# Online Reinforcement Learning for Dynamic Multimedia Systems

Nicholas Mastronarde*, Mihaela van der Schaar

## ABSTRACT

In our previous work, we proposed a systematic cross-layer framework for dynamic multimedia systems, which allows each layer to make *autonomous* and *foresighted* decisions that maximize the system's *long-term performance*, while meeting the application's real-time delay constraints. The proposed solution solved the cross-layer optimization *offline*, under the assumption that the multimedia system's probabilistic dynamics were *known* a priori, by modeling the system as a layered Markov decision process. In practice, however, these dynamics are *unknown* a priori and therefore must be learned *online*. In this paper, we address this problem by allowing the multimedia system layers to learn, through repeated interactions with each other, to autonomously optimize the system's long-term performance at run-time. The two key challenges in this multi-agent (layered) learning setting are: (i) each layer's learning performance is directly impacted by not only its own dynamics, but also by the learning processes of the other layers with which it interacts; and, (ii) selecting a learning model that appropriately balances time-complexity (i.e. learning speed) with the multimedia system's limited memory and the multimedia application's real-time delay constraints. We propose two reinforcement learning algorithms for optimizing the system under different design constraints: the first algorithm solves the cross-layer optimization in a centralized manner, and the second solves it in a decentralized manner. We analyze both algorithms in terms of their required computation, memory, and inter-layer communication overheads. After noting that the proposed reinforcement learning algorithms learn too slowly, we introduce a complementary accelerated learning algorithm that exploits partial knowledge about the system's dynamics in order to dramatically improve the system's performance. In our experiments, we demonstrate that decentralized learning can perform as well as centralized learning, while enabling the layers to act autonomously. Additionally, we show that existing application-independent reinforcement learning algorithms, and existing myopic learning algorithms deployed in multimedia systems, perform significantly worse than our proposed application-aware and foresighted learning methods.

## I. INTRODUCTION

State-of-the-art multimedia technology is poised to enable widespread proliferation of a variety of life-enhancing applications, such as video conferencing, emergency services, surveillance, telemedicine, remote teaching and training, augmented reality, and distributed gaming. However, efficiently designing and implementing such delay-sensitive multimedia applications on resource-constrained, heterogeneous devices and systems is challenging due to the real-time constraints, high workload complexity, and the time-varying environmental dynamics experienced by the system (e.g. video source characteristics, user requirements, workload characteristics, number of running applications, memory/cache behavior etc.).

Cross-layer adaptation is an increasingly popular solution for addressing these challenges in dynamic multimedia systems (DMSs) [1]-[8]. This is because the performance of DMSs (e.g. video rate-distortion costs, delay, and power consumption) can be significantly improved by jointly optimizing parameters, configurations, and algorithms across two or more system layers (i.e. the application, operating system, and hardware layers), rather than optimizing and adapting them in isolation. However, existing cross-layer solutions have several important limitations.

*Centralized cross-layer solutions*: The majority of these solutions require a central optimizer to coordinate the application, operating system, and hardware adaptations to optimize the performance of one or multiple multimedia applications sharing the DMS's limited resources. To this end, a new interface between the central optimizer and all of the layers is created, requiring extensive modifications to the operating system [4] [5] or the introduction of an entirely new middleware layer [1] [2] [3] [24] [25]. Unfortunately, these approaches ignore the fact that the majority of modern system's are comprised of layers (components or modules) that are designed by different manufacturers, which makes such tightly integrated designs impractical, if not impossible [8] [18] [19].

*Myopic cross-layer solutions*: Another limitation of many existing cross-layer solutions for DMSs is that they are *myopic*. In other words, cross-layer decisions are made reactively in order to optimize the *immediate* utility, without considering the impact of these decisions on the future utility. However, in DMSs, it is essential to predict the impact of the current decisions on the long-term utility because the multimedia source characteristics, workload characteristics, OS and hardware dynamics are often correlated across time. Moreover, in DMSs, utility fluctuations across time lead to poor user experience and bad resource planning leads to inefficient resource usage and wasted power [26] [27]. In contrast, *foresighted* (i.e. long-term) optimization techniques take into account the impact of immediate cross-layer decisions on the DMS's expected future utility. Importantly, foresighted policy optimizations have been successfully deployed at the hardware layer to solve the dynamic power management problem [11]-[13]; however, with the exception of our prior work [9], they have never been used for cross-layer DMS optimization, where *all the layers make foresighted decisions*.

*Single-layer learning solutions*: The various multimedia system layers must be able to adapt at run-time to their experienced dynamics. Most existing learning solutions for multimedia systems are concerned with modeling the unknown and potentially time-varying environment experienced by a single layer. We refer to these as *single-agent* learning solutions (in our setting, an agent corresponds to a layer).

A broad range of single-agent techniques have been deployed in multimedia (and general purpose) systems in recent years, which include estimation techniques such as maximum likelihood estimation [4] [5] [10] [12] [13], statistical fitting [7], regression methods [28], adaptive linear prediction [29], and ad-hoc estimation heuristics [3]. However, these solutions ignore the fact that DMSs do not only experience dynamics at a single layer (e.g. time-varying workload), but at multiple (possibly all) layers, which interact with each other. Hence, it is necessary to enable the layers to learn at run-time how to autonomously and asynchronously adapt their processing strategies based on forecasts about (i) their *future dynamics* and, importantly, (ii) *how they impact and are impacted by the other layers*. For this reason, we will model the run-time cross-layer optimization as a cooperative multi-agent learning problem [30] (where the layers are the agents).

In summary, while significant contributions have been made to enhance the performance of DMSs using cross-layer design techniques, no systematic cross-layer framework exists that explicitly considers: (i) the design and implementation constraints imposed by a layered DMS architecture; (ii) the ability of the various layers to autonomously make foresighted decisions in order to jointly maximize the DMS's utility; and, most importantly, (iii) how layers can learn their unknown environmental dynamics and determine how they impact and are impacted by the other layers.

In this paper, similar to [6] [7] [8], we consider a multimedia system in which (i) a video encoder at the application layer makes rate-distortion-complexity tradeoffs by adapting its configuration and (ii) the operating system layer makes energy-delay tradeoffs by adapting the hardware layer's operating frequency. Our contributions are as follows:

- We propose a centralized reinforcement learning algorithm for online cross-layer DMS optimization based on the well-known Q-learning algorithm. This centralized solution can be easily implemented if a single manufacturer designs all of the layers in the multimedia system.

- We propose a novel multi-agent (layered) Q-learning algorithm, which allows the multimedia system layers to autonomously learn their optimal foresighted policies online, through repeated interactions with each other. This decentralized solution can be implemented if the layers are designed by different manufacturers. We show experimentally that the proposed layered learning algorithm, which adheres to the layered system architecture, performs as well as a traditional centralized learning algorithm, which does not.

- We propose a reinforcement learning technique, which is complementary to the aforementioned learning algorithms, that exploits partial a priori knowledge about the system's dynamics in order to accelerate the rate of learning and improve overall learning performance. Unlike existing reinforcement learning techniques [14], [15], which can only learn about previously visited state-action pairs, the proposed algorithm exploits our partial knowledge about the system's dynamics in order to learn about multiple state-action pairs even before they have been visited.

The remainder of this paper is organized as follows. In Section II, we present the cross-layer problem formulation. In Section III, we describe the two layer multimedia system model. In Section IV, we propose a centralized and layered (decentralized) Q-learning algorithm for run-time cross-layer optimization. In Section V, we propose a technique to accelerate the rate of learning by exploiting partial a priori knowledge that we have about the system's dynamics. In Section VI, we present our experimental results and we conclude the paper in Section VII.

## II. CROSS-LAYER PROBLEM FORMULATION

In this section, we define the layered structure of the system under study and formulate the cross-layer system optimization problem as a Markov decision process (MDP). In Section III, we discuss in detail how the considered DMS problem can be modeled within the proposed cross-layer MDP framework.

### A. Layered system model

We model the layered system as a tuple $\Omega = \langle \mathcal{L}, \mathcal{A}, \mathcal{S}, p, R \rangle$, where

- $\mathcal{L} = \{1, ..., L\}$ is a set of $L$ autonomous layers, which participate[1] in the cross-layer optimization. Each layer is indexed $l \in \{1, ..., L\}$ with layer 1 corresponding to the lowest participating layer (e.g. the HW layer) and layer $L$ corresponding to the highest participating layer (e.g. the APP layer).

- $\mathcal{A}$ is the global action set $\mathcal{A} = \times_{l \in \mathcal{L}} \mathcal{A}_l$[2], where $\mathcal{A}_l$ is the $l$ th layer's local action set (e.g. the APP layer's available source-coding parameter configurations and the OS layer's available commands for switching the CPU's operating frequency).

- $\mathcal{S}$ is the global state set $\mathcal{S} = \times_{l \in \mathcal{L}} \mathcal{S}_l$, where $\mathcal{S}_l$ is the $l$ th layer's local state set (e.g. the set of buffer states at the APP layer).

- $p$ is the joint transition probability function mapping the global state, global action, and global next-state to a value in $[0,1]$, i.e. $p : \mathcal{S} \times \mathcal{A} \times \mathcal{S} \mapsto [0,1]$.

- $R$ is the expected reward function, which maps the global state and global action to a real number representing the system's expected reward, i.e. $R : \mathcal{S} \times \mathcal{A} \mapsto \mathbb{R}_+$.

Within the proposed DMS framework, we are interested in trading off multiple objectives including application performance (characterized by delay and video distortion) and power consumption. Hence, we

---

[1] We note that for a layer to "participate" in the cross-layer optimization it must be able to adapt one or more of its parameters, configurations, or algorithms (e.g. the APP layer can adapt its source coding parameters); alternatively, if a layer does not "participate", then it is omitted.

[2] $\times_{l \in \mathcal{L}} \mathcal{A}_l = \mathcal{A}_1 \times \cdots \times \mathcal{A}_L$ is the $L$-ary Cartesian product.

assume that the reward function is of the form:

$$R(\boldsymbol{s}, \boldsymbol{a}) = g_L(\boldsymbol{s}, a_L) - \sum_{l \in \mathcal{L}} \omega_l J_l(s_l, a_l), \tag{1}$$

where $J_l$ is the local cost at layer $l$, which is independent of the states and actions at the other layers (e.g. the application's distortion and hardware's power consumption); the utility gain $g_L(\boldsymbol{s}, a_L)$ is related to the delay experienced by the application at layer $L$ given its own state and action, and the states of the other system layers, which support it (see Section III.C for details); and, $\omega_l$ weights the relative importance of the layer costs with the utility gain. In Section III, we formulate an illustrative DMS problem within this layered modeling framework.

## B. System optimization objective

Unlike existing cross-layer optimization solutions, which focus on optimizing the myopic (i.e. immediate) utility, the goal in the proposed cross-layer framework is to find the optimal actions at each stage $n \in \mathbb{N}$ that maximize the *discounted sum of future rewards* [9] [11] [14] [15], i.e.

$$\sum_{n=0}^{\infty} (\gamma)^n R(\boldsymbol{s}^n, \boldsymbol{a}^n \mid \boldsymbol{s}^0) \tag{2}$$

where the parameter $\gamma$ $(0 \leq \gamma < 1)$ is the "discount factor," which defines the relative importance of present and future rewards, and $\boldsymbol{s}^0$ is the initial state. We refer to the Markov decision policy $\pi^* : \mathcal{S} \mapsto \mathcal{A}$, which maximizes the discounted sum of future rewards from each initial state $\boldsymbol{s}^0 \in \mathcal{S}$, as the optimal *foresighted* policy. We use a *discounted* sum of rewards instead of, for example, the *average* reward, because: (i) typical video sources have temporally correlated statistics over short time intervals such that the future environmental dynamics cannot be easily predicted without error [7], and therefore the system may benefit by weighting its immediate reward more heavily than future rewards; and, (ii) The multimedia session's lifetime is not known a priori (i.e. the session may end unexpectedly) and therefore rewards should be maximized sooner rather than later.

Throughout this paper, we will find it convenient to work with the optimal *action-value function* $Q^* : \mathcal{S} \times \mathcal{A} \mapsto \mathbb{R}$ [15]:

$$Q^*(\boldsymbol{s}, \boldsymbol{a}) = R(\boldsymbol{s}, \boldsymbol{a}) + \gamma \sum_{\boldsymbol{s}' \in \mathcal{S}} p(\boldsymbol{s}' \mid \boldsymbol{s}, \boldsymbol{a}) V^*(\boldsymbol{s}'), \tag{3}$$

where $V^*(\boldsymbol{s}) = \max_{\boldsymbol{a}} Q^*(\boldsymbol{s}, \boldsymbol{a})$, $\forall \boldsymbol{s} \in \mathcal{S}$, is the optimal *state-value function* [15]. In words, $Q^*(\boldsymbol{s}, \boldsymbol{a})$ is the expected sum of discounted rewards achieved by taking action $\boldsymbol{a}$ in state $\boldsymbol{s}$ and then following the optimal policy $\pi^* : \mathcal{S} \mapsto \mathcal{A}$ thereafter, where

$$\pi^*(\boldsymbol{s}) = \arg\max_{\boldsymbol{a}} Q^*(\boldsymbol{s}, \boldsymbol{a}), \ \forall \boldsymbol{s} \in \mathcal{S}. \tag{4}$$

## III. ILLUSTRATIVE SYSTEM SPECIFICATION

In this section, using the framework introduced above, we model a system in which two layers participate in the cross-layer optimization (i.e. $\mathcal{L} = \{1, 2\}$). In particular, the APP layer makes rate-distortion-complexity tradeoffs by adapting its configuration and the OS layer makes energy-delay tradeoffs by adapting the HW layer's operating frequency (using, for example, the platform independent open standard known as the Advanced Configuration and Power Interface [11]). Due to the fact that the OS layer controls the HW layer, we combine them into a single decision making layer, which we call the joint OS/HW layer.

As in [21], we model the video source as a sequence of *video data units* (for example, video macroblocks, groups of macroblocks, or pictures), which arrive at a constant rate into the application's pre-encoding buffer. Similar to [11], we assume that the system operates over discrete time slots. Additionally, we assume that the time slots have variable length, such that one data unit is encoded in each time slot. We interchangeably refer to the time slot during which the $n$ th data unit is encoded as the $n$ th "time slot" or "stage," where $n \in \mathbb{N}$.

For clarity, instead of specifying the layers by their indices, in this section we use the subscripts APP to represent quantities related to the APP layer ($l = 2$) and OS to represent quantities related to the joint OS/HW layer's parameters ($l = 1$). Fig. 1 illustrates the system model, which we describe in detail in the following subsections, and Table I provides an abbreviated list of the notation introduced in this section.

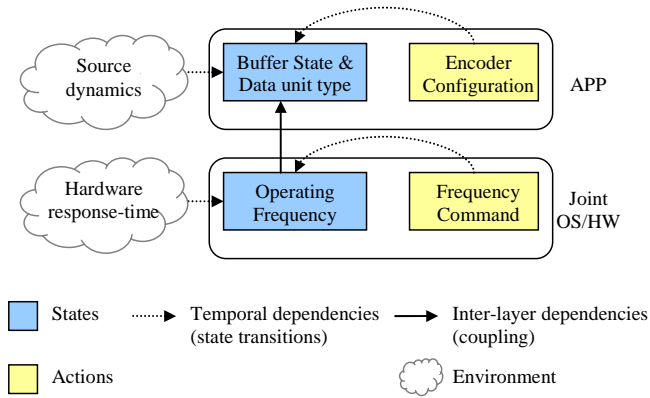

Fig. 1. System model diagram showing the states, actions, and environment at each layer. Although not illustrated here, the actions at each layer are selected based on the global system state.

Table I. Abbreviated list of notation

| $f$ | Operating frequency | $u$ | Frequency command |
|---|---|---|---|
| $z$ | Data unit type | $h$ | Encoding parameter configuration |
| $q$ | Buffer occupancy | $a_1 = a_{\text{OS}} = u$ | Joint OS/HW layer action |
| $s_1 = s_{\text{OS}} = f$ | Joint OS/HW layer state | $a_2 = a_{\text{APP}} = h$ | APP layer action |
| $s_2 = s_{\text{APP}} = (z, q)$ | APP layer state | $\boldsymbol{a} = (a_1, a_2)$ | Global action |
| $\boldsymbol{s} = (s_1, s_2)$ | Global state | $g_2(\boldsymbol{s}, a_2)$ | Utility gain |
| $p_1(s_1' \mid s_1, a_1),$ $p_2(s_2' \mid \boldsymbol{s}, a_2)$ | Transition probability functions for the joint OS/HW and APP layers | $J_1(s_1, a_1),$ $J_2(s_2, a_2)$ | Cost functions for the joint OS/HW and APP layers |

## A. Hardware Layer Specification (HW Layer)

The $n$ th data unit is processed at the HW layer's current operating frequency $f^n$, which is a member of the joint OS/HW layer's state set $\mathcal{S}_{\text{OS}} = \{f_i : i = 1, \ldots, N_f\}$. We let the joint OS/HW layer's cost represent the power dissipated at operating frequency $f^n \in \mathcal{S}_{\text{OS}}$, i.e. $J_{\text{OS}}(f^n) = P(f^n)$ (watts), where $P(\cdot)$ is the system's power-frequency function (see [5] [7] for example power-frequency functions).

At stage $n$, the OS layer can issue a command in its action set $\mathcal{A}_{\text{OS}} = \{u : u \in \mathcal{S}_{\text{OS}}\}$ to change the HW layer's operating frequency. Similar to [11], we assume that there is a non-deterministic delay associated with the operating frequency transition such that

$$p_{\text{OS}}^{(f)}\left(F^{n+1} = f^{n+1} \mid F^n = f^n, u^n\right) = \begin{cases} \beta, & \text{if } f^{n+1} = u^n \\ 1 - \beta, & \text{if } f^{n+1} = f^n \\ 0, & \text{otherwise} \end{cases} \tag{5}$$

where $\{F^n\} = \{F^n : n \in \mathbb{N}\}$ is the sequence of operating frequencies; $f^n$ and $f^{n+1}$ are the operating frequencies at stage $n$ and $n+1$, respectively; and, $\beta \in (0,1]$ is the probability of a successful operating frequency transition (with probability $1 - \beta$ the operating frequency remains the same). In Fig. 1, this transition is illustrated by a temporal dependency between the frequency command (action) and operating frequency (state) in the joint OS/HW layer. Note that, regardless of the OS layer's action $u^n$, the $n$ th data unit is processed at the operating frequency $f^n \in \mathcal{S}_{\text{OS}}$ at the power-cost $J_{\text{OS}}(f^n)$.

## B. Application Layer Specification (APP Layer)

The $n$ th data unit can be classified as being one of $N_z$ types in the state set $\mathcal{S}_{\text{APP}}^{\langle z \rangle} = \{z_i : i = 1, \ldots, N_z\}$. The set of states $\mathcal{S}_{\text{APP}}^{\langle z \rangle}$ depends on the specific video coder being used at the APP layer. In this paper, for illustration, we assume that $N_z = 3$ because video streams are typically compressed into group of pictures structures containing intra-predicted (I), inter-predicted (P), and bi-directionally predicted (B) data units (e.g. MPEG-2, MPEG-4, and H.264/AVC); however, the set of data unit types can be further refined based on, for example, each frame's activity level [7] [20]. It has been shown that transitions among data unit types and frame activity levels in an (adaptive) group of pictures structure can be modeled as a stationary Markov process [7] [20] [31]: i.e.,

$$p_{\mathrm{APP}}^{(z)}(Z^{n+1} = z^{n+1} \mid Z^n = z^n) \qquad (6)$$

where $\{Z^n\} = \{Z^n : n \in \mathbb{N}\}$ is the sequence of data unit types, and $z^n$ and $z^{n+1}$ are the types of the data units that are encoded in time slots $n$ and $n+1$, respectively. We assume that the choice of data unit type is governed by an algorithm similar to the one in [20]; therefore, the probabilities in (6) depend on the desired ratio of I, P, and B data units (e.g. IBBP…), the source characteristics (i.e. the APP layer's environment illustrated in Fig. 1), and the condition used to decide when to code an I frame [20].

The $n$ th data unit can be coded using any one of $N_h$ encoding parameter configurations in the APP layer's action set $\mathcal{A}_{\mathrm{APP}} = \{h_i : i = 1, \ldots, N_h\}$ (for example, a video encoder can adapt at run-time its choice of quantization parameter, macroblock partition size, deblocking filter, entropy coding scheme, sub-pixel motion accuracy, motion-vector search range, and motion estimation search algorithm). Given the data unit type $z^n \in \mathcal{S}_{\mathrm{APP}}^{(z)}$, each configuration $h^n \in \mathcal{A}_{\mathrm{APP}}$ achieves different operating points in the rate-distortion-complexity space [6]. We let $b^n(z^n, h^n)$, $d^n(z^n, h^n)$, and $c^n(z^n, h^n)$ represent the encoded bit-rate (bits per data unit), encoded distortion (mean square error), and encoding complexity (cycles), respectively. We assume that $b^n(z^n, h^n)$, $d^n(z^n, h^n)$, and $c^n(z^n, h^n)$ are instances of the i.i.d. random variables $B(z^n, h^n)$, $D(z^n, h^n)$, and $C(z^n, h^n)$, respectively. Note that we do not need to know the distributions of these random variables to learn the optimal state-value function and optimal policy at run-time (see Section IV).

We penalize the APP layer's action $h^n \in \mathcal{A}_{\mathrm{APP}}$ by employing the Lagrangian cost measure used in the H.264/AVC reference encoder for making rate-distortion optimal mode decisions: i.e. we define the APP layer's cost as $J_{\mathrm{APP}}(z^n, h^n) = d^n(z^n, h^n) + \lambda_{rd} b^n(z^n, h^n)$, where $d^n(z^n, h^n)$ is the $n$ th data unit's encoded distortion; $b^n(z^n, h^n)$ is its encoded bit-rate; and $\lambda_{rd} \in [0, \infty)$ is a Lagrangian multiplier, which can be set based on the rate-constraints.

At stage $n$, the APP layer's pre-encoding buffer contains $q^n$ data units, where $q^n$ is a member of the APP layer's pre-encoding buffer state set $\mathcal{S}_{\mathrm{APP}}^{(q)} = \{q_i : i = 0, \ldots, N_q\}$ and $N_q$ is the maximum number of data units that can be stored in the buffer. In this paper, we assume that $N_q$ is not limited by the available memory, but instead it is determined by the specific application's delay-constraints [21]. We assume that data units arrive in the pre-encoding buffer at $\eta$ data units per second based on a deterministic arrival process from the video capture device (for example, if data units are frames, then $\eta$ will typically be 15, 24, or 30 frames per second). If the buffer is full when a data unit arrives, then that data unit is discarded.

The pre-encoding buffer's state at stage $n+1$ can be expressed recursively based on its state at stage $n$ : i.e.,

$$q^{n+1} = \min\left\{\left[q^n + \lfloor t^n(z^n, h^n, f^n) \cdot \eta\rfloor - 1\right]^+, N_q\right\} \tag{7}$$
$$q^0 = q_{\text{init}},$$

where

$$t^n(z^n, h^n, f^n) = \frac{c^n(z^n, h^n)}{f^n} \ \text{(seconds)} \tag{8}$$

is the $n$ th data unit's processing delay, which depends on its complexity $c^n(z^n, h^n)$ and the processor's operating frequency $f^n$ (in Fig. 1, the coupling between the operating frequency and processing delay is represented by the inter-layer dependency arrow); $\lfloor x\rfloor$ is the integer part of $x$ ; and $\lfloor x\rfloor^+ = \max\{x, 0\}$ . Note that $c^n(z^n, h^n)$ is an instance of the random variable $C(z^n, h^n)$ with distribution $C(z^n, h^n) \sim p_C\left(C = c^n \mid z^n, h^n\right)$ .

Recall that the system operates over discrete time slots of variable length, such that one data unit is encoded in each time slot. Hence, in (7), the $-1$ indicates that the $n$ th data unit departs the pre-encoding buffer after the non-deterministic encoding delay $T(f^n, z^n, h^n) = C(z^n, h^n)/f^n$ (seconds). Meanwhile, the number of data units that arrive in the pre-encoding buffer during the $n$ th time slot, $\tilde{t}^n(z^n, h^n, f^n) = t^n(z^n, h^n, f^n) \cdot \eta$ , is an instance of the random variable $\tilde{T}(f^n, z^n, h^n) = \eta C(z^n, h^n)/f^n$ (data units) with distribution

$$\tilde{T}(f^n, z^n, h^n) \sim p_{\tilde{T}}(\tilde{T} = \tilde{t}^n \mid f^n, z^n, h^n) = \frac{f^n}{\eta} \cdot p_C\left(\frac{f^n}{\eta} \cdot c^n \mid z^n, h^n\right) \tag{9}$$

Based on (7) and (9), the pre-encoding buffer's state transition can be modeled as a controllable Markov chain with transition probabilities

$$p_{\text{APP}}^{(q)}\left(Q^{n+1} = q^{n+1} \mid Q^n = q^n, f^n, z^n, h^n\right)$$
$$= \begin{cases} p_{\tilde{T}}\left\{\tilde{T} < 2 \mid f^n, z^n, h^n\right\}, & q^n = q^{n+1} = 0 \\ p_{\tilde{T}}\left\{\tilde{T} \geq N_q - q + 1 \mid f^n, z^n, h^n\right\}, & q^{n+1} = N_q \\ p_{\tilde{T}}\left\{q^{n+1} - q^n + 1 \leq \tilde{T} < q^{n+1} - q^n + 2 \mid f^n, z^n, h^n\right\}, & \text{otherwise,} \end{cases} \tag{10}$$

where $\{Q^n\} = \{Q^n : n \in \mathbb{N}\}$ is the sequence of buffer states; $\{\tilde{T}^n\} = \{\tilde{T}^n : n \in \mathbb{N}\}$ is the sequence of data unit arrivals; and, $p_{\text{APP}}^{(q)}\left(Q^{n+1} = q^{n+1} \mid Q^n = q^n, f^n, z^n, h^n\right)$ is the probability that the queue's occupancy transitions from $q^n \in \mathcal{S}_{\text{APP}}^{(q)}$ to $q^{n+1} \in \mathcal{S}_{\text{APP}}^{(q)}$ given the operating frequency $f^n \in \mathcal{S}_{\text{OS}}$, the data unit type $z^n \in \mathcal{S}_{\text{APP}}^{(z)}$, and the APP layer's configuration $h^n \in \mathcal{A}_{\text{APP}}$. Note that, from (7), $q^{n+1} - q^n \geq -1$, with equality when there are no data unit arrivals, i.e. $\tilde{t}^n = 0$.

Noting that the transitions of the data unit's type (see (6)) and the pre-encoding buffer's state (see (10)) are conditionally independent given the current data unit type $z^n$, the APP layer's transition probability function can be expressed as

$$
\begin{aligned}
&p_{\text{APP}}\left(\boldsymbol{s}_{\text{APP}}^{n+1} = (z^{n+1}, q^{n+1}) \mid \boldsymbol{s}_{\text{APP}}^n = (z^n, q^n), f^n, h^n\right) \\
&\quad = p_{\text{APP}}^{(q)}\left(Q^{n+1} = q^{n+1} \mid Q^n = q^n, f^n, z^n, h^n\right) \cdot p_{\text{APP}}^{(z)}\left(Z^{n+1} = z^{n+1} \mid Z^n = z^n\right)
\end{aligned}, \quad (11)
$$

where $\boldsymbol{s}_{\text{APP}}^n = (z^n, q^n) \in \mathcal{S}_{\text{APP}}$ and $\boldsymbol{s}_{\text{APP}}^{n+1} = (z^{n+1}, q^{n+1})$ are the APP layer's state at time slot $n$ and $n+1$, respectively. Meanwhile, the joint OS/HW layer's transition probability function can be expressed as (see (5))

$$
p_{\text{OS}}\left(s_{\text{OS}}^{n+1} = f^{n+1} \mid s_{\text{OS}}^n = f^n, u^n\right) = p_{\text{OS}}^{(f)}\left(F^{n+1} = f^{n+1} \mid F^n = f^n, u^n\right). \quad (12)
$$

Because the state transitions at the APP layer and joint OS/HW layer are independent, the global transition probability function defined in Section II.A can be factored as

$$
\begin{aligned}
p(\boldsymbol{s}^{n+1} \mid \boldsymbol{s}^n, \boldsymbol{a}^n) = {}& p_{\text{OS}}\left(s_{\text{OS}}^{n+1} = f^{n+1} \mid s_{\text{OS}}^n = f^n, u^n\right) \cdot \\
& p_{\text{APP}}\left(\boldsymbol{s}_{\text{APP}}^{n+1} = (z^{n+1}, q^{n+1}) \mid \boldsymbol{s}_{\text{APP}}^n = (z^n, q^n), f^n, h^n\right),
\end{aligned} \quad (13)
$$

where $\boldsymbol{s}$ and $\boldsymbol{a}$ are defined as in Table I.

## C. System Reward Details

Recall the definition of the reward function in (1). Thus far, we have defined the costs at each layer, but we have not defined the utility gain $g_{\text{APP}}(\boldsymbol{s}^n, a_2^n)$. As mentioned in Section II.A, the utility gain is closely related to the application's experienced queuing delay in the pre-encoding buffer. Conventionally, multimedia buffers are used to enforce a maximum tolerable processing [22] or transmission delay [21] and as long as the buffer does not overflow (i.e. the overflow constraint is not violated), there is no penalty. In our setting, however, the multimedia system's dynamics are unknown and non-stationary; therefore, we cannot guarantee that the buffer overflow constraints will be satisfied. In light of this, we design the utility gain to non-linearly reward the system for maintaining queuing delays less than the maximum tolerable delay, thereby protecting against overflows that may result from a sudden increase in encoding delay. Formally, we define the utility gain at stage $n$ as

$$
g_{\text{APP}}(\boldsymbol{s}^n, a_2^n) = 1 - \left(\frac{q^n + \lfloor \tilde{t}^n \rfloor - 1}{N_q}\right)^2, \quad (14)
$$

which is near its maximum when the buffer is empty (corresponding to zero queuing delay) and is minimized when the buffer is full. Note that $\tilde{t}^n = \tilde{t}^n(f^n, z^n, h^n)$, but we omit the arguments for brevity. In Appendix A, we discuss in detail why (14) is a good definition for the utility gain.

Given the utility gain defined in (14), the APP layer's cost $J_{\text{APP}}(z^n, h^n)$ defined in Section III.B, and the joint OS/HW layer's cost $J_{\text{OS}}(f^n)$ defined in Section III.A, the reward at stage $n$ defined in (1) can be rewritten as

$$R(\boldsymbol{s}^n, \boldsymbol{a}^n) = \underbrace{1 - \left( \frac{q^n + \lfloor \tilde{t}^n \rfloor - 1}{N_q} \right)^2}_{g_{\text{APP}}(\boldsymbol{s}^n, a_2^n)} - \omega_{\text{OS}} \underbrace{P(f^n)}_{J_{\text{OS}}(f^n)} - \omega_{\text{APP}} \underbrace{[d^n(z^n, h^n) + \lambda_{rd} b^n(z^n, h^n)]}_{J_{\text{APP}}(z^n, h^n)}. \quad (15)$$

# IV. Learning the Optimal Decision Policy

## A. Selecting a Learning Model

As we mentioned before, the multimedia system's dynamics (i.e. $R(\boldsymbol{s}, \boldsymbol{a})$ and $p(\boldsymbol{s}' \mid \boldsymbol{s}, \boldsymbol{a})$) are *unknown* and therefore the optimal action-value function $Q^*$ and the optimal policy $\pi^*$ must be learned online, based on experience. However, it remains to explain how we can learn the optimal policy online despite these unknown quantities. Let us consider the following two options:

- *Direct estimation and policy computation:* One option is to directly estimate the reward and transition probability function using, for example, maximum likelihood estimation, and then to perform value iteration[3] [15] to determine the optimal policy; however, this solution is far too complex because each iteration of the algorithm has complexity $O\left( |\mathcal{S}|^2 |\mathcal{A}| \right)$ and the number of iterations required to converge to the optimal policy is polynomial in the discount factor $\gamma$. Also note that storing the estimated transition probability function incurs a memory overhead of $O\left( |\mathcal{S}|^2 |\mathcal{A}| \right)$, which for a large number of states is unacceptably large (e.g. our experimental test-bed in Section VI uses 765 states and 15 actions, and therefore requires over 33 MBs of memory to directly store the transition probability function using 32 bit single-precision floating point numbers).

- *Offline policy computation and online policy interpolation (supervised learning):* To avoid using the complex value iteration algorithm online, in [12] the authors propose a supervised learning approach involving an offline stage and an online stage. Offline, they quantize every unknown probability parameter into $NS$ samples (more samples yield more accurate results). For each quantized probability bin (acting as a training sample), they compute a corresponding policy (acting as a desired output). Then, online, they directly estimate the transition probability function and, based on this, interpolate the offline computed policies to determine the current policy. For $M$ parameters, this learning algorithm requires $\prod_{m=1}^{M} NS_m$ policy look-up tables of size $|\boldsymbol{S} \times \boldsymbol{A}|$ as defined in [11]. Thus, the memory overheads for this solution are $O\left(|\boldsymbol{S} \times \boldsymbol{A}| \prod_{m=1}^{M} NS_m\right)$, which, even for modest values of $M$ and $NS$, requires much more memory than the direct estimation solution above.

Clearly, given the multimedia system's limited memory and the multimedia application's real-time delay constraints, neither of the above solutions are practical. Thus, in our resource-constrained cross-layer setting with many states and actions (and many unknown parameters), we adopt a *model-free* reinforcement learning solution, which can be used to learn the optimal action-value function $Q^*$ and optimal policy $\pi^*$ online, without directly estimating the reward and transition probability functions. Specifically, we adopt a low complexity algorithm called Q-learning, which also has limited memory requirements (see Table III in Section IV.D). In our cross-layer setting, the conventional Q-learning algorithm can be implemented by a central optimizer -- located at either the APP layer, the joint OS/HW layer, or a separate middleware layer.

### B. Proposed Centralized Q-learning

Central to the conventional centralized Q-learning algorithm is a simple update step performed at the end of each time slot based on the *experience tuple* (ET) $\left(\boldsymbol{s}^n, \boldsymbol{a}^n, r^n, \boldsymbol{s}^{n+1}\right)$:

$$\delta^n = \left[r^n + \gamma \max_{\boldsymbol{a}' \in \boldsymbol{A}} Q^n(\boldsymbol{s}^{n+1}, \boldsymbol{a}')\right] - Q^n(\boldsymbol{s}^n, \boldsymbol{a}^n), \tag{16}$$

$$Q^{n+1}(\boldsymbol{s}^n, \boldsymbol{a}^n) \leftarrow Q^n(\boldsymbol{s}^n, \boldsymbol{a}^n) + \alpha^n \delta^n, \tag{17}$$

where $\boldsymbol{s}^n$, $\boldsymbol{a}^n$, and $r^n = g_L^n - \sum_{l \in \boldsymbol{\mathcal{L}}} \omega_l J_l^n$ are the state, performed action, and corresponding reward in time slot $n$, respectively; $\boldsymbol{s}^{n+1}$ is the resulting state in time slot $n+1$; $\boldsymbol{a}'$ is the *greedy action*[4] in state $\boldsymbol{s}^{n+1}$; $\delta^n$ is the so-called *temporal-difference* (TD) error [14]; and, $\alpha^n \in [0,1]$ is a time-varying learning rate parameter. We note that the action-value function can be initialized arbitrarily at time $n = 0$.

It is well known that if (i) the rewards and transition probability functions are stationary, (ii) all of the state-action pairs are visited infinitely often, and (iii) $\alpha^n$ satisfies the *stochastic approximation*

---

[3] Policy iteration or linear programming could also be used, but these solutions are also too complex.

[4] A greedy action $\boldsymbol{a}^*$ is one that maximizes the current estimate of the action-value function, i.e. $\boldsymbol{a}^* = \arg\max_{\boldsymbol{a}} \left\{Q(\boldsymbol{s}, \boldsymbol{a})\right\}$.

*conditions*[5] $\sum_{n=0}^{\infty} (\alpha^n) = \infty$ and $\sum_{n=0}^{\infty} (\alpha^n)^2 < \infty$, then $Q$ converges with probability 1 to $Q^*$ [16]. Subsequently, the optimal policy can be found using (4).

During the learning process, it is not entirely obvious what the best action is to take in each state. On the one hand, the optimal action-value function can be learned by randomly *exploring* the available actions in each state. Unfortunately, unguided randomized exploration cannot guarantee acceptable performance during the learning process. On the other hand, taking greedy actions, which *exploit* the available information in the action-value function $Q(\boldsymbol{s}, \boldsymbol{a})$, can guarantee a certain level of performance. Unfortunately, exploiting what is already known about the system prevents the discovery of new, better, actions. To judiciously trade off exploration and exploitation, we use the so-called $\varepsilon$-*greedy* action selection method [14]:

$\varepsilon$-*greedy* **action selection:** With probability $1 - \varepsilon$, take the greedy action that maximizes the action-value function, i.e. $\boldsymbol{a}^* = \arg\max_{\boldsymbol{a}} \{Q(\boldsymbol{s}, \boldsymbol{a})\}$; and, with probability $\varepsilon$, take an action randomly and uniformly over the action set.

We write $\boldsymbol{a} = \Phi_\varepsilon(Q, \boldsymbol{s})$ to show that $\boldsymbol{a}$ is an $\varepsilon$-*greedy* joint-action. Note that, if the $\varepsilon$-*greedy* action is taken from the optimal state-value function $Q^*$, then actions corresponding to the optimal policy will only be taken with probability $1 - \varepsilon$; therefore, if $\varepsilon$ does not decay to 0, then optimal performance will not be achieved.

Fig. 2 illustrates the information exchanges required to deploy the centralized Q-learning algorithm in a two layer multimedia system during one time slot. In Fig. 2, the top and bottom blocks represent the APP layer and joint OS/HW layer, respectively. The center block represents the centralized optimizer, which selects both layers' actions and updates the system's global action-value function $Q$. As we mentioned before, the centralized optimizer may be located at either the APP layer, the joint OS/HW layer, or a separate middleware layer. To best highlight the information exchanges, we illustrate the centralized optimizer as a separate middleware layer. Although we do not explicitly indicate this in Fig. 2, the ET $(\boldsymbol{s}, \boldsymbol{a}, r, \boldsymbol{s}')$ is used by the block labeled "Update $Q(\boldsymbol{s}, \boldsymbol{a})$," which performs the update step defined in (17).

---

[5] For example, $\alpha^n = 1/(n+1)$ satisfies the stochastic approximation conditions.

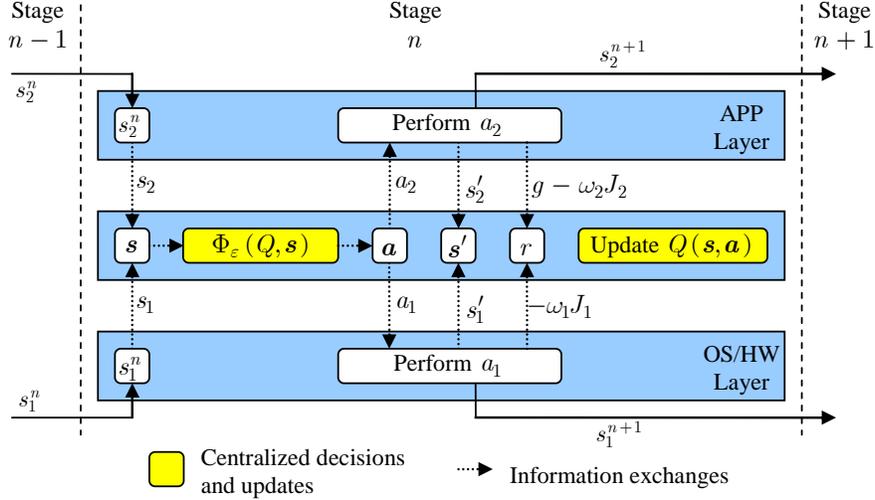

Fig. 2. Information exchanges required to deploy the centralized Q-learning update step in one time slot.

### C. Proposed Layered Q-learning

Thus far, we have discussed how the traditional Q-learning algorithm can be implemented by a centralized optimizer. As we discussed in the introduction, this centralized solution is most appropriate when a single manufacturer designs all of the system's layers. However, we are also interested in developing learning solutions that allow layers designed by different manufactures to act autonomously, but still cooperatively optimize the system's performance. This is a challenging problem in our *cross-layer* setting because the dynamics experienced at each layer may depend on the learning processes of the other layers. Hence, a key challenge is determining how to coordinate the autonomous learning processes of the various layers to enable them to successfully learn. In this section, we propose a *layered Q-learning* algorithm, which we derive by exploiting the structure of the transition probability and reward functions in order to decompose the global action-value function into local action-value functions at each layer. Then, in subsection IV.D, we compare the computation, communication, and memory overheads incurred by the centralized and the proposed layered learning algorithms.

In this subsection, we first show how the action-value function can be decomposed by exploiting the structure of the considered cross-layer problem. Subsequently, we show how this decomposition leads to a layered Q-learning algorithm, which solves the cross-layer optimization online, in a *decentralized manner*, when the transition probability and reward functions are unknown a priori. In this section, for notational simplicity, we drop the time slot index $n$, and denote the next-state $s^{n+1}$ using $s'$.

#### 1) Action-Value Function Decomposition

The proposed decomposition assumes that (i) each layer has access to the global state $s$ in each time slot; (ii) the APP layer knows the number of states at the joint OS/HW layer, i.e. $|\mathcal{S}_{\text{OS}}|$; and, (iii) the joint OS/HW layer knows the number of states and actions at the APP layer, i.e. $|\mathcal{S}_{\text{APP}}|$ and $|\mathcal{A}_{\text{APP}}|$.

Given the additive reward function defined in (15) and the factored transition probability function defined in (13), we can rewrite the optimal action-value function defined in (3) as follows:

$$Q^*(\boldsymbol{s}, \boldsymbol{a}) = g_2(\boldsymbol{s}, a_2) - \omega_1 J_1(s_1, a_1) - \omega_2 J_2(s_2, a_2) + \\ \gamma \sum_{s_1' \in \mathcal{S}_1, s_2' \in \mathcal{S}_2} p_1(s_1' \mid s_1, a_1) p_2(s_2' \mid s_2, a_2) V^*(\boldsymbol{s}'),$$

(18)

where, $V^*(\boldsymbol{s}') = \max_{\boldsymbol{a}' \in \mathcal{A}} Q^*(\boldsymbol{s}', \boldsymbol{a}')$ is the optimal state-value for state $\boldsymbol{s}' = (s_1', s_2')$ (refer to Table I for a review of the notation).

Observing that $g_2(\boldsymbol{s}, a_2) - \omega_2 J_2(s_2, a_2)$ is independent of $s_1'$ and that $\sum_{s_1' \in \mathcal{S}_1} p_1(s_1' \mid s_1, a_1) = 1$, we may rewrite (18) as follows:

$$Q^*(\boldsymbol{s}, a_1, a_2) = -\omega_1 J_1(s_1, a_1) + \\ \sum_{s_1' \in \mathcal{S}_1} \left[ p_1(s_1' \mid s_1, a_1) \left( \begin{array}{l} g_2(\boldsymbol{s}, a_2) - \omega_2 J_2(s_2, a_2) + \\ \gamma \sum_{s_2' \in \mathcal{S}_2} p_2(s_2' \mid s_2, a_2) V^*(s_1', s_2') \end{array} \right) \right].$$

(19)

Given the global state $\boldsymbol{s} = (s_1, s_2)$ and the optimal state-value function $V^*$, the APP layer can perform the inner computation:

$$Q_1^*(\boldsymbol{s}, a_2, s_1') = \left( \begin{array}{l} g_2(\boldsymbol{s}, a_2) - \omega_2 J_2(s_2, a_2) + \\ \gamma \sum_{s_2' \in \mathcal{S}_2} p_2(s_2' \mid s_2, a_2) V^*(s_1', s_2') \end{array} \right), \ \forall a_2 \in \mathcal{A}_2 \text{ and } \forall s_1' \in \mathcal{S}_1$$

(20)

which is independent of the immediate action at the joint OS/HW layer (i.e. $a_1$), but depends on the joint OS/HW layer's potential next-state $s_1'$. For each global state $\boldsymbol{s} = (s_1, s_2)$, the APP layer must compute the set $\left\{ Q_1^*(\boldsymbol{s}, a_2, s_1') : a_2 \in \mathcal{A}_2, s_1' \in \mathcal{S}_1 \right\}$ using (20). Then, given this set from the APP layer, the joint OS/HW layer can perform the outer computation in (19): i.e.,

$$Q^*(\boldsymbol{s}, a_1, a_2) = \left( \begin{array}{l} -\omega_1 J_1(s_1, a_1) + \\ \sum_{s_1' \in \mathcal{S}_1} p_1(s_1' \mid s_1, a_1) Q_1^*(\boldsymbol{s}, a_2, s_1') \end{array} \right), \ \forall a_1 \in \mathcal{A}_1 \text{ and } \forall a_2 \in \mathcal{A}_2$$

(21)

$Q^*$ can be computed by repeating this procedure for all $\boldsymbol{s} \in \mathcal{S}$.

Intuitively, $Q_1^*(\boldsymbol{s}, a_2, s_1')$ can be interpreted as the APP layer's estimate of the expected discounted future rewards; however, there are two important differences between $Q_1^*(\boldsymbol{s}, a_2, s_1')$ and the centralized action-value function $Q^*(\boldsymbol{s}, \boldsymbol{a})$. First, $Q_1^*(\boldsymbol{s}, a_2, s_1')$ does not include the immediate reward at the joint OS/HW layer (i.e. $-\omega_1 J_1(s_1, a_1)$) because it is unknown to the APP layer; and, second, $Q_1^*(\boldsymbol{s}, a_2, s_1')$ depends on the joint OS/HW layer's next-state $s_1'$ (instead of the expectation over of the next-states as it does in the centralized case) because the APP layer does not know the joint OS/HW layer's transition probability function. After receiving $Q_1^*\left(\boldsymbol{s}, a_2, s_1'\right)$ from the APP layer, the joint OS/HW layer is able to "fill in" this missing information. Clearly, from the derivation, $Q^*\left(\boldsymbol{s}, a_1, a_2\right)$ at the joint OS/HW layer is equivalent to the centralized action-value function $Q^*(\boldsymbol{s}, \boldsymbol{a})$.

For more information about this decomposition, we refer the interested reader to our prior work [9], in which we use a similar decomposition to solve the cross-layer optimization *offline*, in a decentralized manner, under the assumption that the transition probability and reward functions are *known* a priori.

### 2) Layered Q-learning

Recall that the centralized Q-learning algorithm described in Section IV.B violates the layered system architecture because it requires a centralized manager to select actions for both of the layers. In contrast, the layered Q-learning algorithm that we propose in this subsection – made possible by the action-value function decomposition described above – adheres to the layered architecture by enabling each layer to autonomously select its own actions and to update its own action-value function. In the following, we discuss the local information requirements for each layer, how each layer selects its local actions, and how each layer updates its local action-value function.

***Local information:*** The proposed layered Q-learning algorithm requires that the APP layer maintains an estimate of the action-value function on the left-hand side of (20): i.e.,

$$\left\{Q_1\left(\boldsymbol{s}, a_2, s_1'\right): \boldsymbol{s} \in \boldsymbol{\mathcal{S}}, a_2 \in \mathcal{A}_2, s_1' \in \mathcal{S}_1\right\},$$

and that the joint OS/HW layer maintains an estimate of the action-value function on the left-hand side of (21): i.e.,

$$\left\{Q\left(\boldsymbol{s}, a_1, a_2\right): \boldsymbol{s} \in \boldsymbol{\mathcal{S}}, (a_1, a_2) \in \boldsymbol{\mathcal{A}}\right\}.$$

At time slot $n = 0$, these tables can be initialized to 0, i.e. $Q_1^0\left(\boldsymbol{s}, a_2, s_1'\right) = 0$ and $Q^0\left(\boldsymbol{s}, a_1, a_2\right) = 0$. Thus, at initialization, the only information required by the APP layer about the joint OS/HW layer is the number of states at the joint OS/HW layer, i.e. $|\mathcal{S}_{\mathrm{OS}}|$; and, the only information required by the joint OS/HW layer about the APP layer is the number of states and actions at the APP layer, i.e. $|\mathcal{S}_{\mathrm{APP}}|$ and $|\mathcal{A}_{\mathrm{APP}}|$.

***Local action selection:*** At run-time, given the current global state $\boldsymbol{s} = (s_1, s_2)$, the APP layer selects an $\varepsilon$-greedy action $a_2 \in \mathcal{A}_2$ as described in Section IV.B, but with the greedy action selected as follows:

$$\left(a_2^*, \hat{s}_1'\right) = \underset{a_2 \in \mathcal{A}_2, \, s_1' \in \mathcal{S}_1}{\arg\max} \left\{Q_1\left(\boldsymbol{s}, a_2, s_1'\right)\right\}, \tag{22}$$

and, the joint OS/HW layer selects its $\varepsilon$-greedy action $a_1 \in \mathcal{A}_1$, but with the greedy action selected as follows:

$$(a_1^*, \hat{a}_2) = \underset{a_1 \in \mathcal{A}_1, a_2 \in \mathcal{A}_2}{\arg \max} \{Q(\boldsymbol{s}, a_1, a_2)\}. \tag{23}$$

Note that the APP layer selects action $a_2^*$ under the assumption that the joint OS/HW layer will select its action $a_1^*$ to transition to the next-state $\hat{s}_1'$, which maximizes the APP layer's *estimated* discounted future rewards $Q_1(\boldsymbol{s}, a_2, s_1')$. Similarly, the joint OS/HW layer selects action $a_1^*$ under the assumption that the APP layer will select action $\hat{a}_2$, which maximizes the joint OS/HW layer's *estimated* discounted future rewards $Q(\boldsymbol{s}, a_1, a_2)$. The message exchanges during the local learning update procedures (described immediately below) serve to "teach" each layer how they impact and are impacted by the other layer. Thus, through the learning process, the layers improve their assumptions about each other, which enables them to improve their greedy actions.

***Local learning updates:*** After executing the $\varepsilon$-greedy action $\boldsymbol{a}^n = (a_1^n, a_2^n)$ in state $\boldsymbol{s}^n = (s_1^n, s_2^n)$, the system obtains the reward $r^n = g_2^n - \omega_1 J_1^n - \omega_2 J_2^n$ and transitions to state $\boldsymbol{s}^{n+1} = (s_1^{n+1}, s_2^{n+1})$. Based on the experience tuple $(\boldsymbol{s}^n, \boldsymbol{a}^n, r^n, \boldsymbol{s}^{n+1})$, each layer updates its action-value function as follows. First, the joint OS/HW layer must forward the scalar $V^n(s_1^{n+1}, s_2^{n+1}) = \underset{a_1' \in \mathcal{A}_1, a_2' \in \mathcal{A}_2}{\max} Q^n(s_1^{n+1}, s_2^{n+1}, a_1', a_2')$ to the APP layer. Using this forwarded information, the APP layer can perform its action-value function update based on the form of (20): i.e.,

$$\delta_2^n = \left[ g_2^n - \omega_2 J_2^n + \gamma V^n(s_1^{n+1}, s_2^{n+1}) \right] - Q_1^n(\boldsymbol{s}^n, a_2^n, s_1^{n+1}), \tag{24}$$

$$Q_1^{n+1}(\boldsymbol{s}^n, a_2^n, s_1^{n+1}) \leftarrow Q_1^n(\boldsymbol{s}^n, a_2^n, s_1^{n+1}) + \alpha_2^n \delta_2^n. \tag{25}$$

Then, given the scalar $Q_1^{n+1}(\boldsymbol{s}^n, a_2^n, s_1^{n+1})$ from the APP layer and the APP layer's selected action $a_2^n$, the joint OS/HW layer can perform its action-value function update based on the form of (21) as follows:

$$\delta_1^n = \left[ -\omega_1 J_1^n + Q_1^{n+1}(\boldsymbol{s}^n, a_2^n, s_1^{n+1}) \right] - Q^n(\boldsymbol{s}^n, a_1^n, a_2^n), \tag{26}$$

$$Q^{n+1}(\boldsymbol{s}^n, a_1^n, a_2^n) \leftarrow Q^n(\boldsymbol{s}^n, a_1^n, a_2^n) + \alpha_1^n \delta_1^n. \tag{27}$$

Note that, by using the layered Q-learning algorithm, the layers do not need to directly share their local rewards (i.e. the local learning updates only require the APP layer to know its local reward $g_2^n - \omega_2 J_2^n$ and for the joint OS/HW layer to know its local cost $\omega_1 J_1^n$). This is because of the coordinating message exchanges during the local learning update process (i.e. $V^n(s_1^{n+1}, s_2^{n+1})$ is forwarded from the joint OS/HW layer to the APP layer and $Q_1^{n+1}(\boldsymbol{s}^n, a_2^n, s_1^{n+1})$ is forwarded from the APP layer to the joint OS/HW layer).

Fig. 3 illustrates the information exchanges required to deploy the layered Q-learning algorithm in a two layer multimedia system during one time slot. Although we do not explicitly indicate this in Fig. 3, the ETs $\left(\boldsymbol{s}, a_2, g_2 - \omega_2 J_2, s_1'\right)$ and $\left(\boldsymbol{s}, a_1, -\omega_1 J_1, s_1'\right)$ are used by the blocks labeled "Update $Q_1\left(\boldsymbol{s}, a_2, s_1'\right)$" at the APP layer and "Update $Q\left(\boldsymbol{s}, a_1, a_2\right)$" at the joint OS/HW layer, respectively.

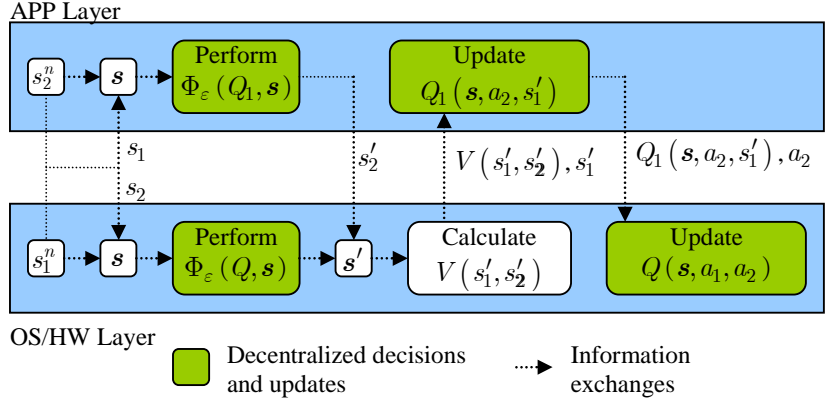

Fig. 3. Information exchanges required to deploy the layered Q-learning update step in one time slot.

## D. Computation, communication, and memory overheads

In this subsection, we compare the computation, communication, and memory overheads associated with the centralized and layered Q-learning algorithms. Fig. 2 and Fig. 3 illustrate the exact information exchanges required for each algorithm; Table II summarizes the greedy action selection procedure and update steps used in each algorithm; and, Table III lists the per time slot computation, communication, and memory overheads associated with each algorithm.

Interestingly, the layered Q-learning algorithm is more complex and requires more memory than the centralized algorithm. Note that, in our setting, $|\mathcal{S}_1| = |\mathcal{A}_1|$; hence, the layered Q-learning algorithm incurs approximately twice the computational and memory overheads as the centralized algorithm. The increased overheads can be interpreted as the cost of optimal decentralized learning (optimal in the sense that it performs as well as the centralized learning algorithm). Nevertheless, the proposed centralized and layered Q-learning algorithms are far less complex and use less memory than the alternative learning solutions discussed in Section IV.A. We also observe that, in both cases, the communication overheads per time slot are $O(1)$ because they are independent of the size of the state and action sets at each layer; however, the precise number of messages exchanged depends on the deployed learning algorithm as illustrated in Fig. 2 and Fig. 3, and noted in Table III.

Table II. Summary of greedy action selection and update steps for various Q-learning algorithms.

| | Greedy Action | Update Step |
|---|---|---|
| Centralized Q-learning | Middleware Layer: $$\boldsymbol{a}^* = \arg\max_{a \in \mathcal{A}} \{Q(\boldsymbol{s}, \boldsymbol{a})\}$$ | Middleware Layer: $$\delta^n = \left[ r^n + \gamma \max_{\boldsymbol{a}' \in \mathcal{A}} Q(\boldsymbol{s}^{n+1}, \boldsymbol{a}') \right] - Q(\boldsymbol{s}^n, \boldsymbol{a}^n)$$ $$Q(\boldsymbol{s}^n, \boldsymbol{a}^n) \leftarrow Q(\boldsymbol{s}^n, \boldsymbol{a}^n) + \alpha^n \delta^n$$ |
| Layered Q-learning | Layer $l = 1$: $$a_1^* = \arg\max_{a_1 \in \mathcal{A}_1, a_2 \in \mathcal{A}_2} \{Q\left(\boldsymbol{s}, a_1, a_2\right)\}$$ Layer $l = 2$: $$a_2^* = \arg\max_{a_2 \in \mathcal{A}_2,\ s_1' \in \mathcal{S}_1} \{Q_1\left(\boldsymbol{s}, a_2, s_1'\right)\}$$ | Layer $l = 1$: $$\delta_1^n = \left\{ \begin{array}{l} \left[ -\omega_1 J_1^n + Q_1^{n+1}\left(\boldsymbol{s}^n, a_2, s_1^{n+1}\right) \right] \\ -Q^n\left(\boldsymbol{s}^n, a_1^n, a_2^n\right) \end{array} \right\}$$ $$Q^{n+1}\left(\boldsymbol{s}^n, a_1^n, a_2^n\right) \leftarrow Q^n\left(\boldsymbol{s}^n, a_1^n, a_2^n\right) + \alpha_1^n \delta_1^n$$ Layer $l = 2$: $$\delta_2^n = \left\{ \begin{array}{l} \left[ g_2^n - \omega_2 J_2^n + \gamma V^n\left(s_1^{n+1}, s_2^{n+1}\right) \right] \\ -Q_1^n\left(\boldsymbol{s}^n, a_2^n, s_1^{n+1}\right) \end{array} \right\}$$ $$Q_1^{n+1}\left(\boldsymbol{s}^n, a_2^n, s_1^{n+1}\right) \leftarrow Q_1^n\left(\boldsymbol{s}^n, a_2^n, s_1^{n+1}\right) + \alpha_2^n \delta_2^n$$ |

Table III. Comparison of computation, memory, and communication overheads (per time slot).

| | Centralized Q-learning | Layered Q-learning |
|---|---|---|
| Computation overheads | Action Selection: $O(\|\mathcal{A}\|)$ Update: $O(\|\mathcal{A}\|)$ | Action Selection: $O(\|\mathcal{A}\| + \|\mathcal{S}_1 \times \mathcal{A}_2\|)$ Update: $O(\|\mathcal{A}\|)$ |
| Memory overheads | $O(\|\boldsymbol{\mathcal{S}} \times \mathcal{A}\|)$ | $O(\|\boldsymbol{\mathcal{S}} \times \mathcal{A}\| + \|\boldsymbol{\mathcal{S}} \times \mathcal{A}_2 \times \mathcal{S}_1\|)$ |
| Communication overheads | $O(1)$ (8 messages) | $O(1)$ (7 messages) |

# V. Accelerated Learning Using Virtual Experience

The large number of buffer states at the APP layer significantly limits the system's learning speed because the Q-learning update step defined in (17) updates the action-value function for only one state-action pair in each time slot. Several existing variants of Q-learning adapt the action-value function for multiple state-action pairs in each time slot. These include model-free temporal-difference-$\lambda$ updates, and model-based algorithms such as Dyna and prioritized sweeping [14] [15]; however, these existing solutions are not system specific and assume no a priori knowledge of the problem's structure. Consequently, they can only update previously visited state-action pairs, leaving the learning algorithm to act blindly (or with very little information) the first few times that it visits each state, thereby resulting in suboptimal learning performance (our experimental results in Section VI.D strongly support this conclusion).

In this subsection, we propose a new Q-learning variant, which is complementary to the abovementioned variants. Unlike conventional reinforcement learning algorithms, which assume that no a priori information is available about the problem's dynamics[6], the proposed algorithm exploits the *form* of the transition probability and reward functions (defined in Section III) in order to update the action-value function for multiple *statistically equivalent*[7] state-action pairs in each time slot, including those that have never been visited. In our specific setting, we exploit the fact that the data unit arrival distribution $p_{\tilde{T}}(\tilde{t} \mid f, z, h)$ (defined in (9)) is conditionally independent of the current buffer state $q^n$ (given $z^n$, $h^n$, and $f^n$). This allows us to extrapolate the experience obtained in each time slot to other buffer states and action pairs. Thus, in stationary (non-stationary) environments, the proposed algorithm improves convergence time (adaptation speed) at the expense of increased computational complexity. For ease of exposition, we discuss the algorithm in terms of the centralized system; however, it can be easily extended to work with the layered Q-learning algorithm proposed in Section IV.C.

Let $\boldsymbol{\sigma}^n = (\boldsymbol{s}^n, \boldsymbol{a}^n, r^n, \boldsymbol{s}^{n+1})$ represent the ET at stage $n$, where $\boldsymbol{s}^n = (z^n, q^n, f^n)$, $\boldsymbol{a}^n = (u^n, h^n)$, $r^n = g_2^n - \omega_1 J_1^n - \omega_2 J_2^n$, and $\boldsymbol{s}^{n+1} = (z^{n+1}, q^{n+1}, f^{n+1})$. Given $q^n$ and the number of data unit arrivals $\lfloor \tilde{t}^n \rfloor = \lfloor \tilde{t}^n(z^n, h^n, f^n) \rfloor$, the next buffer state $q^{n+1}$ can be easily determined from (7). By exploiting our partial knowledge about the system's dynamics, we can use the statistical information provided by $\lfloor \tilde{t}^n \rfloor$ to generate *virtual experience tuples* (virtual ETs) that are statistically equivalent to the actual ET. This statistical equivalence allows us to perform the Q-learning update step for the virtual ETs using information provided by the actual ET.

We let $\tilde{\boldsymbol{\sigma}}^n = (\tilde{\boldsymbol{s}}, \tilde{\boldsymbol{a}}, \tilde{r}, \tilde{\boldsymbol{s}}') \in \Sigma(\boldsymbol{\sigma}^n)$ represent one virtual ET in the set of virtual ETs $\Sigma(\boldsymbol{\sigma}^n)$. In order to be statistically equivalent to the actual ET, the virtual ETs in $\Sigma(\boldsymbol{\sigma}^n)$ must satisfy the following two conditions:

1. The data unit arrival distribution $p_{\tilde{T}}(\tilde{t} \mid f, z, h)$ defined in (9) must be the same for the virtual ETs as it is for the actual ET. In other words, the virtual operating frequency $\tilde{f}$, the virtual type $\tilde{z}$, and the virtual configuration $\tilde{h}$ must be the same as the actual operating frequency $f^n$, the actual type $z^n$, and the actual configuration $h^n$, respectively. This also implies that the virtual costs at the APP and joint OS/HW layers are the same as the actual costs at these layers, i.e. $\tilde{J}_1 = J_1^n$ and $\tilde{J}_2 = J_2^n$.

2. The next virtual buffer state $\tilde{q}' \in \mathcal{S}_{\mathrm{APP}}^{(q)}$ must be related to the current virtual buffer state $\tilde{q} \in \mathcal{S}_{\mathrm{APP}}^{(q)}$ through the buffer evolution equation defined in (7): i.e.,

$$\tilde{q}' = \min\left\{\left[\tilde{q} + \lfloor \tilde{t}^n \rfloor - 1\right]^+, N_q\right\}, \tag{28}$$

where $\lfloor \tilde{t}^n \rfloor$ is the number of data unit arrivals under the actual ET.

Any virtual ET that satisfies the two above conditions can have its reward determined using information embedded in the actual ET. Specifically, from the first condition, we know that the virtual ET's local costs are $\tilde{J}_1 = J_1^n$ and $\tilde{J}_2 = J_2^n$. Then, based on the second condition, we can compute the virtual ET's utility gain as $\tilde{g}_2 = 1 - \left(\dfrac{\tilde{q} + \lfloor \tilde{t}^n \rfloor - 1}{N_q}\right)^2$. Finally, the Q-learning update step defined in (17) can be performed on every virtual ET $\tilde{\boldsymbol{\sigma}}^n = \left(\tilde{\boldsymbol{s}}, \tilde{\boldsymbol{a}}, \tilde{r}, \tilde{\boldsymbol{s}}'\right) \in \boldsymbol{\Sigma}(\boldsymbol{\sigma}^n)$ as if it is the actual ET.

Performing the Q-learning update step on every virtual ET in $\boldsymbol{\Sigma}(\boldsymbol{\sigma}^n)$ incurs a computational overhead of approximately $O\left(|\boldsymbol{\Sigma}(\boldsymbol{\sigma}^n) \times \boldsymbol{\mathcal{A}}|\right)$ in time slot $n$ (note that $\boldsymbol{\Sigma}(\boldsymbol{\sigma}^n)$ is typically as large as the buffer state set $\mathcal{S}_{\mathrm{APP}}^{(q)} = \left\{q : i = 0, \ldots, N_q\right\}$ if you include the actual ET). Unfortunately, it may be impractical to incur such large overheads in every time slot, especially if the data unit granularity is small (e.g. one macroblock). Hence, in our experimental results in Section VI, we show how the learning performance is impacted by updating $\Psi \le |\boldsymbol{\Sigma}(\boldsymbol{\sigma}^n)|$ virtual ETs in each time slot by selecting them randomly and uniformly from the virtual ET set $\boldsymbol{\Sigma}(\boldsymbol{\sigma}^n)$.

Table IV describes the virtual ET based learning procedure in pseudo-code and Fig. 4 compares the backup diagram [14] of the conventional Q-learning algorithm to the proposed algorithm with virtual ET updates. Importantly, because the virtual ETs are statistically equivalent to the actual ET, the virtual ET based learning algorithm is *not* an approximation of the Q-learning algorithm; in fact, the proposed algorithm accelerates Q-learning by exploiting our partial knowledge about the structure of the considered problem.

Table IV. Accelerated learning using virtual experience tuples.

| 1. | **Initialize** $Q(s, a)$ arbitrarily for all $(s, a) \in \mathcal{S} \times \mathcal{A}$ ; |
|---|---|
| 2. | **Initialize** state $s^0$ ; |
| 3. | **For** $n = 0, 1, \dots$ |
| 4. | Take action $a^n$ using $\varepsilon$-greedy action selection on $Q(s^n, \cdot)$ ; |
| 5. | Obtain experience tuple $\sigma^n = (s^n, a^n, r^n, s^{n+1})$ and record $\lfloor \tilde{t}^n \rfloor$ ; |
| 6. | **For** all $\tilde{q} \in \mathcal{S}_{\text{APP}}^{(q)}$      % Generate the virtual ET set $\Sigma(\sigma^n)$ |
| 7. | $\tilde{q}' = \min \left\{ \left[ \tilde{q} + \lfloor \tilde{t}^n \rfloor - 1 \right]^+, N_q \right\}$ ;    % virtual ET next buffer state |
| 8. | $\tilde{s} = (\tilde{f}, \tilde{z}, \tilde{q}) = (f^n, z^n, \tilde{q})$ ;     % virtual ET state |
| 9. | $\tilde{a} = (\tilde{u}, \tilde{h}) = (u^n, h^n)$ ;     % virtual ET action |
| 10. | $\tilde{r} = \left[ 1 - \left( \frac{\tilde{q} + \lfloor \tilde{t} \rfloor - 1}{N_q} \right)^2 \right] - \omega_1 J_1^n - \omega_2 J_2^n$ ;    % virtual ET reward |
| 11. | $\tilde{s}' = (\tilde{f}', \tilde{z}', \tilde{q}') = (f^{n+1}, z^{n+1}, \tilde{q}')$ ;    % virtual ET next state |
| 12. | $\tilde{\sigma} = (\tilde{s}, \tilde{a}, \tilde{r}, \tilde{s}') \in \Sigma(\sigma^n)$ ;    % store virtual ET |
| 13. | **For** $\Psi$ virtual ETs $\tilde{\sigma} \in \Sigma(\sigma^n)$    % Update virtual ETs |
| 14. | $\tilde{\delta} = \left[ \tilde{r} + \gamma \max_{a' \in \mathcal{A}} Q(\tilde{s}', a') \right] - Q(\tilde{s}, \tilde{a})$ ;    % TD error for virtual ET $\tilde{\sigma}^n$ |
| 15. | $Q(\tilde{s}, \tilde{a}) \leftarrow Q(\tilde{s}, \tilde{a}) + \alpha \tilde{\delta}$ ;    % Q-learning update for virtual ET $\tilde{\sigma}^n$ |

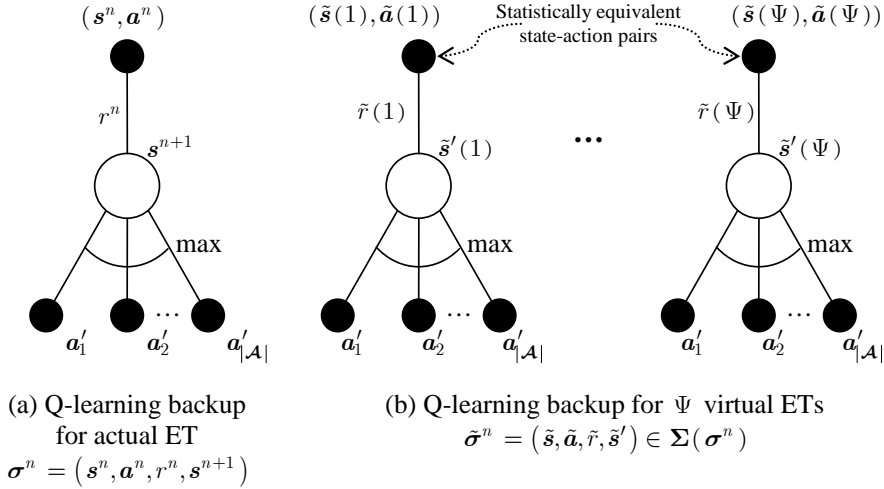

(a) Q-learning backup
for actual ET
$\sigma^n = (s^n, a^n, r^n, s^{n+1})$

(b) Q-learning backup for $\Psi$ virtual ETs
$\tilde{\sigma}^n = (\tilde{s}, \tilde{a}, \tilde{r}, \tilde{s}') \in \Sigma(\sigma^n)$

Fig. 4. Backup diagrams. (a) Conventional Q-learning; (b) Q-learning with virtual updates. The virtual update algorithm applies a backup on the actual ET and an additional $\Psi$ backups on virtual ETs in $\Sigma(\sigma^n)$ indexed by $1 \le \psi \le \Psi$ .

## VI. EXPERIMENTS

In this section, we compare the performance of the proposed learning algorithms against the performance of several existing algorithms in the literature using the cross-layer DMS described in Section III. Table V details the parameters used in our DMS simulator, which we implemented in Matlab. In our simulations, we use actual video encoder trace data, which we obtained by profiling the H.264 JM Reference Encoder (version 13.2) on a Dell Pentium IV computer. Our traces comprise measurements of the encoded bit-rate (bits/MB), reconstructed distortion (MSE), and encoding complexity (cycles) for each video MB of the *Foreman* sequence (30 Hz, CIF resolution, quantization parameter 24) under three different encoding configurations. The chosen parameters are listed in Table V. We use a data unit granularity of one macroblock. As in [7], we assume that the power frequency function is of the form $P(f) = \kappa f^{\theta}$, where $\kappa \in \mathbb{R}_{+}$ and $\theta \in [1,3]$. Since real-time encoding is not possible with the available encoder, we set the data unit arrival rate to $\eta = 44$ DUs/sec, which corresponds to 1/9 frames per second.

For the simulations in Sections VI.B and VI.C, we use stationary data traces, which we generate from the measured (non-stationary) data traces by assuming that the rate, distortion, and complexity samples are drawn from i.i.d. random variables with distributions equivalent to the distributions of the quantities over the entire non-stationary data traces. In Section VI.D, we perform simulations using both the measured (non-stationary) and generated (stationary) data traces.

Table V. Simulation parameters (*Foreman* sequence, 30 Hz, CIF resolution, quantization parameter 24).

| Layer | Parameter | Value |
|---|---|---|
| Application Layer (APP) | Data Unit Granularity | 1 Macroblock |
| | Buffer State Set | $\mathcal{S}_{\mathrm{APP}}^{(q)} = \{0,\ldots,Q\}$, $N_q = 50$ DUs |
| | Data Unit Type Set | $\mathcal{S}_{\mathrm{APP}}^{(z)} = \{z_1, z_2, z_3\}$ $z_1 = \mathrm{P}$, $z_2 = \mathrm{B}$, $z_3 = \mathrm{I}$ |
| | Parameter Configuration | $\mathcal{A}_{\mathrm{APP}} = \{h_1, h_2, h_3\}$ $h_1$ : Quarter-pel MV, 8x8 block ME $h_2$ : Full-pel MV, 8x8 block ME $h_3$ : Full-pel MV, 16x16 block ME |
| | APP Cost Weight | $\omega_{\mathrm{APP}} = 22/1875$ |
| | Rate-Distortion Lagrangian | $\lambda_{\mathrm{rd}} = 1/16$ |
| | Data Unit Arrival Rate | $\eta = 44$ (Data Units/Sec) |
| Operating System / Hardware Layer (Joint OS/HW) | Power-Frequency Function | $P(f) = \kappa f^{\theta}$ $\kappa \in 1.5 \times 10^{-27}$ and $\theta = 3$. |
| | Operating Frequency Set | $\mathcal{A}_{\mathrm{OS}} = \{200, 400, 600, 800, 1000\}$ Mhz |
| | Joint OS/HW Cost Weight | $\omega_{\mathrm{OS}} = 22/125$ |

## A. Evaluation Metrics

We deploy two different metrics to compare the performance of the various learning algorithms discussed in this paper. The first metric is the *weighted estimation error*, which we define as

$$\text{Weighted estimation error} \triangleq \sum_{\boldsymbol{s} \in \mathcal{S}} \boldsymbol{\mu}^*(\boldsymbol{s}) \left| \frac{V^*(\boldsymbol{s}) - V^n(\boldsymbol{s})}{V^*(\boldsymbol{s})} \right|, \qquad (29)$$

where $\boldsymbol{\mu}^*$ is the stationary distribution under the optimal policy $\pi^*$, $V^*(\boldsymbol{s}) = \max_{\boldsymbol{a}} Q^*(\boldsymbol{s}, \boldsymbol{a})$ is the optimal state-value function, and $V^n(\boldsymbol{s}) = \max_{\boldsymbol{a}} Q^n(\boldsymbol{s}, \boldsymbol{a})$ is the state-value function estimate at stage $n$. In (29), we weight the estimation error $|V^*(\boldsymbol{s}) - V^n(\boldsymbol{s})|$ by $\boldsymbol{\mu}^*(\boldsymbol{s})$ because the optimal policy will often only visit a small subset of the states, and the remaining states will never be visited; thus, if we were to estimate $V^*(\boldsymbol{s})$ while following the optimal policy, a *non-weighted* estimation error would not converge to 0, but the proposed weighted estimation error would.

Although the weighted estimation error tells us *how quickly an algorithm can learn the optimal policy*, it does not tell us *how well the algorithm performs while learning*. To see why this is, consider an (exaggerated) scenario in which $|V^*(\boldsymbol{s}) - V^n(\boldsymbol{s})| = 0$, but the exploration rate in the greedy action selection procedure is $\varepsilon = 1$ (i.e. always explore). In this scenario, the greedy actions are optimal, but the random policy implemented by always exploring will be far from optimal. For this reason, we define a second evaluation metric using the *average reward*, which more accurately measures the system's performance. For a simulation of duration $N$, the average reward metric is defined as

$$\text{Average reward} \triangleq \frac{1}{N} \sum_{n=0}^{N-1} r^n, \tag{30}$$

where $r^n$ is the reward sample obtained at stage $n$. If the exploration rate $\varepsilon$ decays appropriately to 0 as $n \to \infty$, then the average reward will increase as the weighted estimation error decreases. Determining an optimal decay strategy for $\varepsilon$ is an important problem, but is beyond the scope of this paper. We refer the interested reader to [32] [14] [15] for more information on the exploitation-exploration trade off.

We note that the weighted estimation error cannot be evaluated under non-stationary dynamics because the optimal state value function cannot be determined. Hence, for non-stationary dynamics, we rely solely on the average reward to evaluate the learning performance. Nevertheless, the rate at which the weighted estimation error converges under stationary dynamics for a particular learning algorithm is indicative of how well that algorithm will perform when the dynamics are non-stationary.

## B. Single-layer Learning Results

Before presenting our cross-layer learning results, we believe it is informative to see how well a single layer can learn (say layer $l$) when the other layer (say layer $-l$) deploys a static policy (see Appendix B for details on how we adapted the Q-learning algorithm to a single-layer setting). For illustration, we assume that the static layer deploys its optimal local policy $\pi^*_{-l}$ corresponding to the global optimal policy $\pi^* = (\pi^*_l, \pi^*_{-l})$; hence, layer $l$ attempts to learn $\pi^*_l$. Fig. 5 illustrates the optimal policies at the APP layer ($\pi^*_2$) and the joint OS/HW layer ($\pi^*_1$) for each buffer state and data unit type when the current operating frequency is $f = 600$ MHz. In Fig. 5, the application actions (i.e. parameter configurations) are as defined in Table VI.

Fig. 6 compares the cumulative average reward obtained using single-layer learning to the optimal achievable reward, and Table VII shows the corresponding power, rate-distortion costs, utility gain, and buffer overflows, for a simulation of duration $N = 192,000$ time slots (approximately 485 frames drawn from the Foreman sequence, CIF resolution, by repeating the sequence from the beginning after 300 frames). We observe that the total average reward obtained when the joint OS/HW layer learns in response to the APP layer's optimal local policy $\pi_2^*$ is lower than when the APP layer learns in response to the joint OS/HW layer's optimal local policy $\pi_1^*$. This is because there are more actions to explore at the joint OS/HW layer (5 compared to 3), and the joint OS/HW layer's policy is more important to the system's overall performance than the APP layer's policy for the chosen simulation parameters (see Table V). For instance, given the action sets defined in Table V, the joint OS/HW layer can significantly impact the application's experienced delay (i.e. a factor of 5 change from 200 MHz to 1000 Mhz), while the APP layer cannot (i.e. its actions impact the delay by less than a factor of 2). Consequently, when the APP layer learns in response to the joint OS/HW layer's optimal policy (Fig. 5(b)), the system is initially better off than when the joint OS/HW layer learns in response to the APP layer's optimal policy (Fig. 5(a)).

We note that, in Fig. 6, the saturated performance of the APP and joint OS/HW layers' best-response learning algorithms is due to the finite exploration probability (i.e. $\varepsilon > 0$ in the $\varepsilon$-greedy action selection procedure), which forces the layers to occasionally test suboptimal actions.

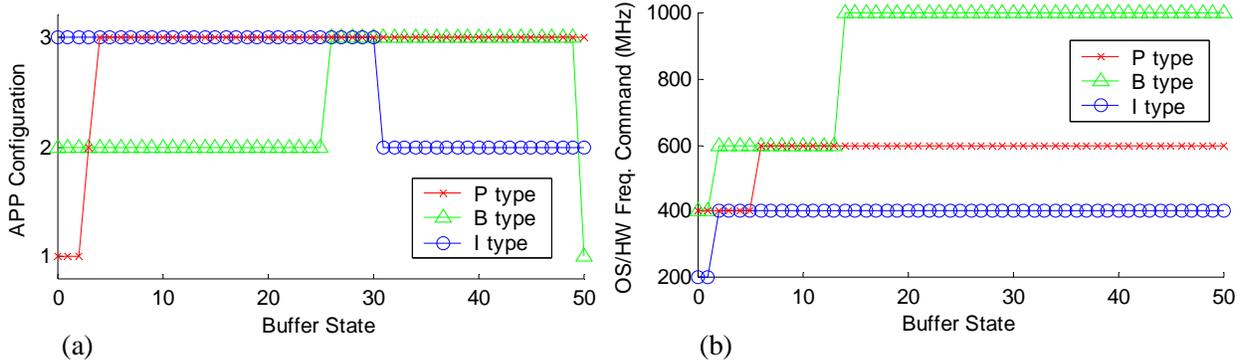

Fig. 5. Optimal policies for each data unit type. (a) APP parameter configurations. (b) Joint OS/HW layer frequency command. The current operating frequency is set to be $f = 600$ Mhz.

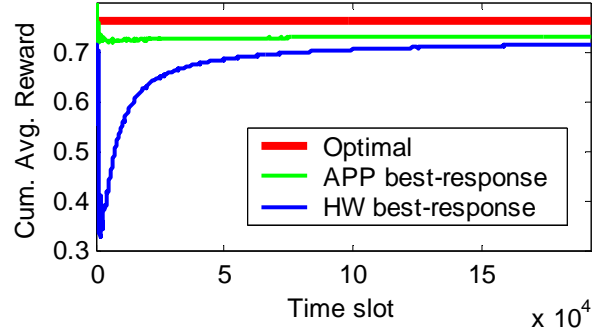

Fig. 6. Cumulative average reward for single-layer learning with stationary trace ( $N = 192,000$ ).

Table VII. Single-layer learning performance statistics for stationary trace ( $N = 192,000$ ).

|  | Best-response APP Layer | Best-response Joint OS/HW Layer | Optimal |
|---|---|---|---|
| Avg. Reward | 0.7337 | 0.7172 | 0.7631 |
| Avg. Power (W) | 0.2835 | 0.4512 | 0.2435 |
| Avg. Rate-Distortion | 14.93 | 15.08 | 14.75 |
| Avg. Utility Gain | 0.9588 | 0.9736 | 0.9790 |
| No. Overflows | 0 | 61 | 0 |

## C. Cross-layer Learning Results

In this subsection, we evaluate the performance of the centralized and layered Q-learning algorithms proposed in Section IV.B and IV.C, respectively. Fig. 7 compares the cumulative average reward obtained using layered Q-learning to the performance of the centralized learning algorithm, the optimal achievable reward, and the performance of the myopic learning algorithm deployed in the state-of-the-art cross-layer coordination framework called GRACE-1[8] [5]; Fig. 8 compares the weighted estimation error for the centralized and layered Q-learning algorithms; and, Table VIII shows the corresponding power, rate-distortion costs, utility gain, and buffer overflows. As in the previous simulations, the simulation duration is $N = 192,000$ time slots. Recall from Table III that the algorithms compared in this subsection have roughly the same computational complexity (note that the complexity of the GRACE-1 algorithm is roughly $O(|\mathcal{A}|)$ in our setting because it requires finding the action that will most closely meet a data unit's deadline).

As expected, the layered Q-learning algorithm performs as well as the centralized Q-learning algorithm, but also adheres to the layered architecture because it allows the layers to act autonomously through decentralized decisions and updates. Meanwhile, the myopic learning algorithm deployed by GRACE-1 performs very poorly in terms of overall reward. This is because, for each data unit, GRACE-1 myopically selects the lowest processing frequency that can successfully encode it before its deadline; consequently, the buffer quickly fills, which makes the system susceptible to delay deadline violations

---

[8] In GRACE-1, the statistical cycle demand (complexity) of a video encoding application is measured using an exponential average of the 95th-percentile complexity of recently completed jobs (processed data units) in a sliding window. Based on the statistical cycle demand, GRACE-1

due to the time-varying workload (this is similar to what happens when using the "conventional gain" function described in Appendix A).

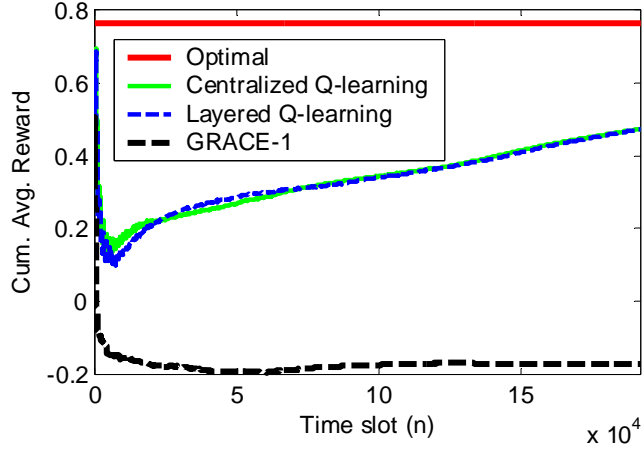

Fig. 7. Cumulative average reward for cross-layer learning algorithms with stationary trace ($N = 192,000$).

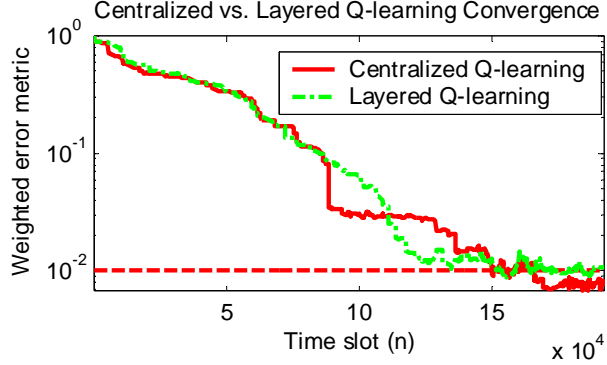

Fig. 8. Weighted estimation error metric for cross-layer learning algorithms with stationary trace ($N = 192,000$).

Table VIII. Cross-layer learning performance statistics for stationary trace ($N = 192,000$).

|  | Optimal | Centralized Q-learning | Layered Q-Learning | GRACE-1 |
|---|---|---|---|---|
| Avg. Reward | 0.7631 | 0.4727 | 0.4721 | -0.1730 |
| Avg. Power (W) | 0.2435 | 0.3815 | 0.4061 | 0.4982 |
| Avg. Rate-Distortion | 14.75 | 15.00 | 14.99 | 15.30 |
| Avg. Utility Gain | 0.9790 | 0.7158 | 0.7195 | 0.0942 |
| No. Overflows | 0 | 1231 | 1251 | 3040 |

It is important to note that, even though the proposed learning algorithms outperform the existing myopic learning algorithm, there is still significant improvement to be made. In particular, the centralized and layered Q-learning algorithms learn too slowly because they only update one state-action pair in each time slot. Consequently, the number of buffer overflows and the power consumption are both higher, and the utility gain is lower, than in the optimal case. In the next subsection, we show that the learning

---

myopically encodes jobs at the lowest frequency which will meet the job's deadline. If slack remains after processing a job, then it is reclaimed for future jobs.

performance (including the weighted estimation error and the average reward) can be dramatically improved by *smartly* updating multiple state-action pairs in each time slot.

## D. Accelerated Learning with Virtual ETs

In this subsection, we compare the performance of the proposed virtual ET based learning algorithm (see Section V) to an existing reinforcement learning algorithm called temporal-difference-$\lambda$ learning[9] (i.e. $TD(\lambda)$ [15]), which has comparable complexity (roughly $O((\Psi+1)|\mathcal{A}|)$ in each time slot for $\Psi$ virtual ET or $TD(\lambda)$ updates in addition to the actual ET update). Recall that, in this subsection, we simulate the system with the measured (non-stationary) trace data and the generated (stationary) trace data.

Fig. 9 illustrates the cumulative average reward achieved when we allow $\Psi \in \{0, 1, 15, 30, 45\}$ virtual ET or $TD(\lambda)$ updates in each time slot (the case when $\Psi = 0$ is equivalent to the conventional centralized Q-learning algorithm in which only the actual ET is updated); Fig. 10 compares the weighted estimation error for the same learning algorithms; and, Table IX illustrates corresponding detailed simulation results.

Interestingly, the $TD(\lambda)$ updates are completely ineffectual at improving the system's performance. In particular, it is obvious that increasing the number of $TD(\lambda)$ updates in each time slot does not guarantee higher average rewards or lower weighted estimation errors. Meanwhile, increasing the number of virtual ET updates in each time slot dramatically improves the weighted estimation error metric and the average reward. The reason for this stark difference in performance is because the virtual ET based learning algorithm can learn about state-action pairs *without visiting them*, while the $TD(\lambda)$ learning algorithm can only learn about *previously visited state-action pairs*. The ability to learn about unvisited states is very important in our problem setting because after the buffer initially fills from early exploration (resulting in the initial drop in rewards illustrated in Fig. 6, Fig. 7, and Fig. 9), the learning algorithm must learn to efficiently drain the buffer without consuming too much power or unnecessarily reducing the application's quality. This is simple when using virtual ETs because the information obtained from experience in a "near full" buffer state (e.g. $q = Q - 1$) can be extrapolated to a "near, near full" buffer state (e.g. $q = Q - 2$) and so on. When using $TD(\lambda)$ updates, however, the algorithm learns very slowly about "near, near full" buffer states because they are visited only infrequently from the "near full" buffer state. For this reason, $TD(\lambda)$ (even for a large number of updates) has trouble emptying the buffer, and performs no better than the conventional centralized Q-learning algorithm, which updates only one state-action pair in each time slot.

One might expect that violating the stationarity assumption, which is widely used in the reinforcement learning literature (see [14][15][16]), would be disastrous to the system's performance. We observe in Fig. 9 and Table IX, however, that the proposed virtual ET based learning algorithm is robust to the non-stationary dynamics in the measured data traces. This is primarily due to its ability to quickly propagate new information obtained at one state-action pair throughout the state-action space. From the data in Table IX, we observe that the percent difference between the reward obtained in the stationary case and the non-stationary case is less than 6%.

As in the single-layer learning case, the learning performance with virtual ETs saturates due to the finite exploration probability (i.e. $\varepsilon > 0$), which prevents the learning algorithm from converging to the optimal policy.

It is important to note that, in these experiments, we have assumed that the update complexity is negligible. This would be true if the updates were performed infrequently (e.g. per video frame) or if the updates were performed in parallel (e.g. using a vector processor); however, it is not a valid assumption for the very frequent and serial updates deployed here (i.e. per video macroblock, without a vector processor). Hence, performing 15, 30, or 45 updates in each time slot is not reasonable, but performing 1 or 2 virtual ET updates is. Despite this technicality, we have shown the relative performance improvements that can be achieved by updating multiple virtual experience tuples in each time slot. An interesting trade off to investigate in future research is the impact of a coarser data unit granularity with a simultaneous increase in the number of virtual updates (such that the average learning complexity remains constant). Lastly, we note that the learning performance could be further improved (for the same number of virtual ET updates) by directing the virtual ET updates to states that are most likely to be visited in the near future (e.g. states near the observed next-state), instead of randomly updating the virtual experience tuples.

---

[9] In our implementation, $TD(\lambda)$ applies the update step defined in (17) to the $\Psi$ most frequently visited states in the recent past. These recently visited states are updated using the TD error obtained for the current state weighted by the previously visited states' "eligibility" as defined in [15]. Informally, the "eligibility" is the discounted frequency with which states have been visited in the past (see [15]).

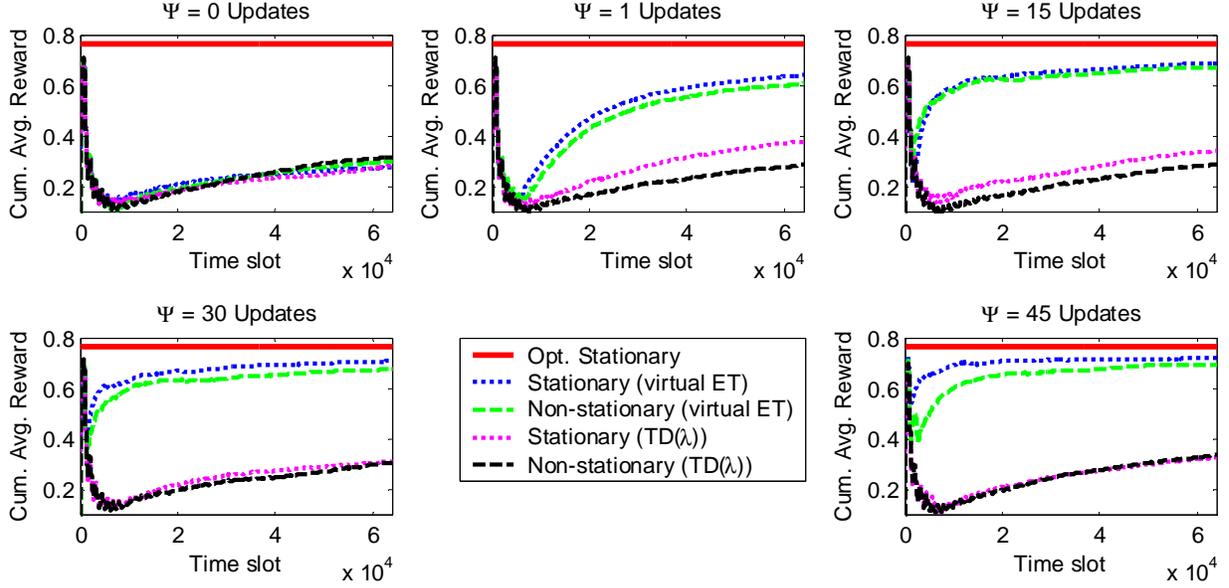

Fig. 9. Cumulative average reward achieved with virtual ET and *TD*($\lambda$) updates for a stationary trace and a non-stationary trace, compared to the optimal achievable reward under the stationary trace ($N = 64,000$).

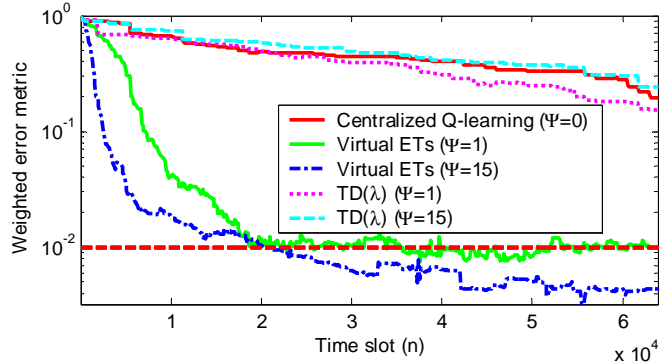

Fig. 10. Weighted estimation error metric for virtual ETs and TD($\lambda$) with stationary trace ($N = 64,000$).

Table IX. Virtual experience learning statistics for stationary (non-stationary) data trace outside (inside) parentheses ($N = 64,000$).

| | Number of Virtual ET Updates | | | | |
|---|---|---|---|---|---|
| | $\Psi = 0$ | $\Psi = 1$ | $\Psi = 15$ | $\Psi = 30$ | $\Psi = 45$ |
| Avg. Reward | 0.2785 (0.3015) | 0.6414 (0.6078) | 0.6893 (0.6710) | 0.7076 (0.6759) | 0.7200 (0.6944) |
| Avg. Power (W) | 0.4212 (0.4632) | 0.4164 (0.4314) | 0.3817 (0.3864) | 0.3432 (0.3657) | 0.3108 (0.3381) |
| Avg. Rate-Distortion | 15.00 (14.89) | 14.92 (14.77) | 15.06 (14.84) | 14.93 (14.81) | 14.87 (14.71) |
| Avg. Utility Gain | 0.5287 (0.5577) | 0.8898 (0.8570) | 0.9332 (0.9131) | 0.9432 (0.9140) | 0.9492 (0.9264) |
| No. Overflows | 1196 (1410) | 1035 (1271) | 502 (437) | 226 (378) | 103 (438) |
| | Number of *TD*($\lambda$) Updates | | | | |
| | $\Psi = 0$ | $\Psi = 1$ | $\Psi = 15$ | $\Psi = 30$ | $\Psi = 45$ |
| Avg. Reward | 0.2835 (0.3188) | 0.3807 (0.2899) | 0.3449 (0.2903) | 0.3141 (0.3088) | 0.3287 (0.3388) |
| Avg. Power (W) | 0.4494 (0.4707) | 0.4649 (0.4732) | 0.4551 (0.4672) | 0.4530 (0.4644) | 0.4544 (0.4875) |
| Avg. Rate-Distortion | 15.07 (14.88) | 15.14 (14.92) | 15.14 (14.88) | 15.03 (14.84) | 15.15 (14.91) |
| Avg. Utility Gain | 0.5395 (0.5763) | 0.6402 (0.5483) | 0.6026 (0.5472) | 0.5703 (0.5647) | 0.5866 (0.5997) |
| No. Overflows | 1255 (1630) | 1230 (1491) | 1267 (1511) | 1234 (1491) | 1250 (1317) |

## VII. Conclusion

Foresighted decisions are required to optimize the performance of resource-constrained multimedia systems. However, in practical settings, in which the system's dynamics are unknown a priori, efficiently learning the optimal foresighted decision policy can be a challenge. Specifically, selecting an improper learning model can lead to an unacceptably slow learning speed, incur excessive memory overheads, and/or adversely impact the application's real-time performance due to large computational overheads. In this paper, we choose reinforcement learning to appropriately balance time-complexity, memory complexity, and computational complexity. We propose a centralized and a layered reinforcement learning algorithm and extend these algorithms to exploit our partial knowledge of the system's dynamics. The proposed accelerated learning algorithm, which is based on updating multiple statistically equivalent state-action pairs in each time slot, allows us to judiciously tradeoff per-stage computational complexity and time-complexity (i.e. learning speed), while incurring low memory overheads.

In our experimental results, we verify that the layered learning solution achieves the same performance as the centralized solution, which may be impractical to implement if the layers are designed by different manufacturers. We also illustrate that our proposed foresighted learning algorithms outperform the myopic learning algorithm deployed in an existing state-of-the-art cross-layer optimization framework and that, by exploiting knowledge about the system's dynamics, we can significantly outperform an existing application-independent reinforcement learning algorithm that has comparable computational overheads.

In the long term, we believe this work can catalyze a shift in the design and implementation of DMSs (and systems in general) by enabling system layers (modules or components) to proactively reason and interact based on forecasts, as well as evolve, and become smarter, by learning from their interactions.

## Appendix A: Utility Gain Function

In this appendix, we discuss the form of the utility gain function defined in (14). Conventionally, if a multimedia buffer does not overflow (i.e. the buffer constraints are not violated), then there is no penalty. Within the proposed MDP-based framework, where we cannot explicitly impose constraints because the dynamics are not known a priori, the conventional buffer model must be integrated into the system's reward function. Specifically, it can be integrated into the reward through a utility gain function of the form:

$$g_{\text{conv}}(\boldsymbol{s}, a_L) = \begin{cases} 1, & q + \lfloor \tilde{t} \rfloor - 1 \leq N_q \\ N_q - (q + \lfloor \tilde{t} \rfloor - 1), & \text{otherwise}, \end{cases}$$

which provides a reward of 1 if the buffer does not overflow, and a penalty proportional to the number of overflows otherwise.

In contrast to the proposed continuous utility gain function defined in (14), $g_{\text{conv}}(\boldsymbol{s}, a_L)$ is disjoint. As a result, the optimal foresighted policy obtained using $g_{\text{conv}}(\boldsymbol{s}, a_L)$ will initially fill the buffer rapidly in an attempt to minimize rate-distortion and power costs (because it is not penalized for filling the buffer) as illustrated in Fig. 11(a). Subsequently, the policy will attempt to keep the buffer nearly full in order to balance the cost of overflow with the rate-distortion and power costs required to reduce the buffer's occupancy. Unfortunately, with the buffer nearly full, any sudden burst in the complexity of a data unit will immediately overflow the buffer as illustrated in Fig. 11(b). In contrast, the proposed utility gain function is robust against bursts in complexity because it encourages the buffer occupancy to remain low as illustrated in Fig. 11(c,d).

The data in Table X shows that the proposed utility gain function not only prevents buffer overflows, but it also achieves comparable power consumption as the conventional utility gain function. Thus, we have verified that our choice of utility gain function is good. An added benefit of the proposed utility gain function is that it aids in the learning process. This is because actions are *immediately* rewarded (or penalized) based on how they impact the buffer state.

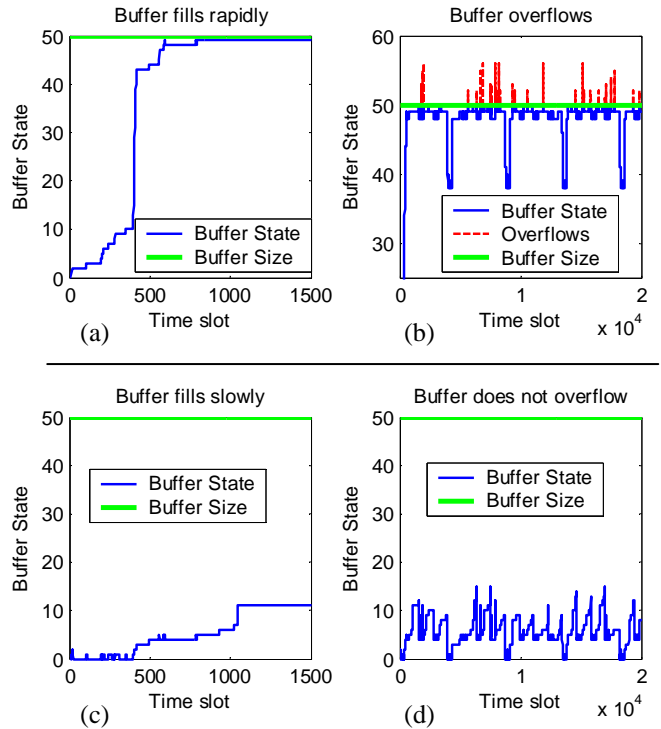

Fig. 11. Buffer evolution in $N = 20,000$ time slot simulation. (a) Conventional utility gain results in the buffer filling rapidly; (b) Conventional utility gain leads to overflows; (c) Proposed utility gain keeps the buffer occupancy low; (d) Proposed utility gain prevents overflows.

Table X. Performance statistics <u>using the conventional utility gain function and the</u> proposed utility gain function.

|  | Form of the utility gain | |
|---|---|---|
|  | Conventional | Proposed |
| Avg. Power (W) | 0.2421 | 0.2427 |
| Avg. Rate-Distortion | 14.95 | 14.84 |
| No. Overflows | 394 | 0 |

# Appendix B: Single-Layer Learning

In the single-layer learning algorithm, the learning layer (say layer $l$) knows the global state $\boldsymbol{s}$, but does not know the actions implemented by the other layer (say layer $-l$), which are dictated by its static policy $\pi_{-l}$. Thus, the stage reward $R(\boldsymbol{s}, \boldsymbol{a})$ defined in (15) that is observed by layer $l$ can be rewritten as

$$R(\boldsymbol{s}, a_l \mid \pi_{-l}) = R(\boldsymbol{s}, (a_l, \pi_{-l}(\boldsymbol{s}))), \tag{31}$$

and the transition probability function defined in (13) can be rewritten as

$$p(\boldsymbol{s}' \mid \boldsymbol{s}, a_l, \pi_{-l}) = p(\boldsymbol{s}' \mid \boldsymbol{s}, (a_l, \pi_{-l}(\boldsymbol{s}))). \tag{32}$$

We define the *local best-response action-value function* at layer $l$ (for a fixed $\pi_{-l}$) as

$$Q_l^*(\boldsymbol{s}, a_l \mid \pi_{-l}) = R(\boldsymbol{s}, a_l \mid \pi_{-l}) + \gamma \sum_{\boldsymbol{s}' \in \mathcal{S}} p(\boldsymbol{s}' \mid \boldsymbol{s}, a_l, \pi_{-l}) \max_{a_l \in \mathcal{A}_l} Q_l^*(\boldsymbol{s}', a_l \mid \pi_{-l}), \tag{33}$$

and the *local best-response policy* at layer $l$, which we denote by $\pi_l^*(\boldsymbol{s} \mid \pi_{-l})$, as

$$\pi_l^*(\boldsymbol{s} \mid \pi_{-l}) = \arg\max_{a_l \in \mathcal{A}_l} Q_l^*(\boldsymbol{s}, a_l \mid \pi_{-l}). \tag{34}$$

Although the local costs at layers $l' \in \mathcal{L}_{-l}$ (i.e. $J_{l'}(s_{l'}, \pi_{l'}(s_{l'}))$) are independent of the $l$th layer's action $a_l$, layer $l$ still needs to know them in order to learn how it impacts the other system layers [17]. For example, if the joint OS/HW layer is unaware of its impact on the application's delay and quality, it will always selfishly minimize its own costs by operating at its lowest frequency and power, which can adversely impact the real-time application's performance. For this reason, in (33), we make the assumption that layer $l$ has access to the global reward.

The $l$th layer's optimal local action-value function $Q_l^*$ (for a fixed $\pi_{-l}$) and the corresponding optimal local best-response policy $\pi_l^*(\boldsymbol{s} \mid \pi_{-l})$ can be learned online using a straightforward adaptation of the centralized Q-learning update step described Section IV.B: i.e.,

$$\delta_l^n = \left[ r^n + \gamma \max_{a_l' \in \mathcal{A}_l} Q_l^n(\boldsymbol{s}^{n+1}, a_l' \mid \pi_{-l}) \right] - Q_l^n(\boldsymbol{s}^n, a_l^n \mid \pi_{-l}), \tag{35}$$

$$Q_l^{n+1}(\boldsymbol{s}^n, a_l^n \mid \pi_{-l}) \leftarrow Q_l^n(\boldsymbol{s}^n, a_l^n \mid \pi_{-l}) + \alpha_l^n \delta_l^n, \tag{36}$$

where $r^n = g_L^n - \sum_{l=1}^{L} \omega_l J_l^n$ is a random sample of the reward with expected value $R(\boldsymbol{s}^n, a_l^n \mid \pi_{-l})$. We refer to this learning procedure at layer $l$ as *local best-response Q-learning*. The best response Q-learning update at layer $l$ requires the experience tuple $\left( s^n, a_l^n, r^n, s^{n+1} \right)$. Note that, given the global state $s^n$, layer $l$ trades off exploitation and exploration by selecting the $\varepsilon$-greedy action $a_l^n = \Phi_\varepsilon \left( Q_l^n, s^n \right)$.